\documentclass{article}

\usepackage{preprint,times}

\usepackage{amsmath,amsfonts,bm}

\def\eqref#1{equation~\ref{#1}}

\def\1{\bm{1}}

\DeclareMathAlphabet{\mathsfit}{\encodingdefault}{\sfdefault}{m}{sl}
\SetMathAlphabet{\mathsfit}{bold}{\encodingdefault}{\sfdefault}{bx}{n}

\newcommand{\E}{\mathbb{E}}

\usepackage{amsmath}
\usepackage{amssymb}
\usepackage{booktabs}
\usepackage{graphicx}
\usepackage{capt-of}
\usepackage{wrapfig}
\usepackage{needspace}
\usepackage{flafter}
\usepackage{microtype}
\usepackage{xcolor}
\usepackage{url}
\usepackage{hyperref}
\hypersetup{
  pdftitle={Beyond Policy Patches: Distilling RL Specialists into Vision-Language-Action Policies},
  pdfauthor={Zhennan Jiang, Tingying Xie, Shilong Ji, Shu'ang Yu, Tianxing Zhou, Yuhui Chen, Jiahuan Gao, Kaizhi Xie, Yuanfan Xu, Feng Gao, Yu Wang, Haoran Li, Dongbin Zhao, Chao Yu},
  colorlinks=true,
  citecolor=blue,
  linkcolor=blue,
  urlcolor=blue
}

\newcommand{\method}{REPAIR}

\title{Beyond Policy Patches: Distilling RL Specialists into Vision-Language-Action Policies}

\author{%
  \begin{minipage}{\textwidth}
  \centering
  \fontsize{10}{14}\selectfont
  Zhennan Jiang\textsuperscript{1,4,5}\quad
  Tingying Xie\textsuperscript{2,6}\quad
  Shilong Ji\textsuperscript{2,3}\quad
  Shu'ang Yu\textsuperscript{2,3}\\
  Tianxing Zhou\textsuperscript{5,6}\quad
  Yuhui Chen\textsuperscript{1,4}\quad
  Jiahuan Gao\textsuperscript{2}\quad
  Kaizhi Xie\textsuperscript{2,3}\quad
  Yuanfan Xu\textsuperscript{2}\\
  Feng Gao\textsuperscript{2}\quad
  Yu Wang\textsuperscript{3}\quad
  Haoran Li\textsuperscript{1,4}\quad
  Dongbin Zhao\textsuperscript{1,4}\quad
  Chao Yu\textsuperscript{3,*}\\[6pt]
  \fontsize{8}{11}\selectfont
  \textsuperscript{1}University of Chinese Academy of Sciences\quad
  \textsuperscript{2}Striding AI\quad
  \textsuperscript{3}Tsinghua University\\
  \textsuperscript{4}Institute of Automation, Chinese Academy of Sciences\quad
  \textsuperscript{5}Zhongguancun Academy\quad
  \textsuperscript{6}Beijing Institute of Technology\\[3pt]
  \textsuperscript{*}Corresponding author
  \end{minipage}%
}
\date{}

\begin{document}

\maketitle

\begin{abstract}
Vision-language-action (VLA) policies can handle noncritical stages of long-horizon manipulation yet fail during short precision-critical phases. 
Lightweight reinforcement learning (RL) specialists can address these failures without directly applying RL to the full VLA, but retain auxiliary controllers and switching mechanisms at deployment. 
We study how to transfer their capabilities into a single flow-matching VLA. 
Direct imitation of specialist rollouts can reproduce action jitter and leave the VLA unprepared for states encountered during its own execution. 
We introduce REPAIR, a value-guided distillation framework that combines offline initialization with on-policy refinement. 
At each critical-phase observation, REPAIR queries the corresponding specialist for multiple action candidates and weights them using its critic to construct a supervision distribution. 
A shared weighted flow-matching objective trains the VLA first on observations from specialist-assisted rollouts, then on observations collected through its own execution. 
Specialists provide supervision without taking control during on-policy collection and are removed at deployment. 
Across six real-robot tasks with 30 evaluation trials per task, REPAIR achieves 100\% success on all six tasks, with lower mean end-effector jerk than specialist-assisted execution on every task.
Additional visual results are available at \url{https://repair-vla.github.io/}.
\end{abstract}

\section{Introduction}
\label{sec:introduction}

Vision-language-action (VLA) policies provide a general foundation for robotic manipulation~\citep{black2024pi0visionlanguageactionflowmodel,intelligence2025pi05visionlanguageactionmodelopenworld,zheng2026x,bjorck2025gr00t}, yet reliable long-horizon execution can be limited by short precision-critical phases~\citep{lim2025longvla}.
A policy may reliably approach an object, yet fail during contact-rich insertion, alignment, fastening, or assembly.
In these phases, millimeter-scale errors, hesitation, and repeated corrections can determine the outcome of the entire episode.
Collecting more human demonstrations is not always sufficient: demonstrations may be inconsistent precisely where the required behavior is most exact, while supervised fine-tuning (SFT) provides no task-reward signal for improving those demonstrated actions.

\begin{figure*}[!t]
  \centering
  \includegraphics[width=\textwidth]{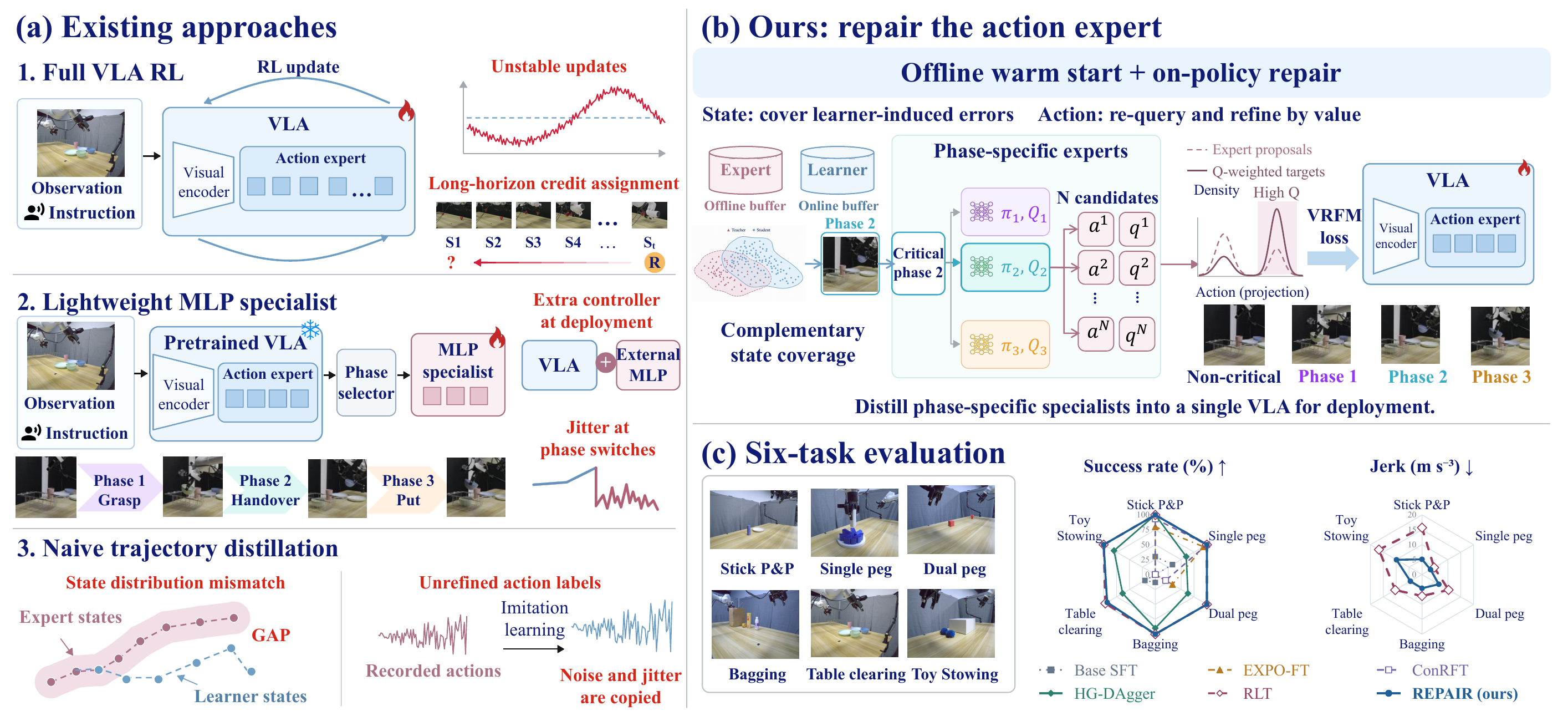}
  \caption{Overview of \method{}. Local RL expertise provides value-guided supervision for offline initialization and on-policy repair. The VLA acts without specialist takeover during online distillation and is deployed as a single policy.}
  \label{fig:overview}
\end{figure*}

Reinforcement learning (RL) offers a way beyond these limitations by improving behavior through interaction and task feedback~\citep{lei2025rl100,chen2026pocp,jiang2025world4rldiffusionworldmodels}.
However, directly applying RL to the VLA's action expert can lead to unstable policy updates~\citep{chen2025conrft,wang2026qvgm}.
Full-task RL introduces a separate difficulty in long-horizon settings, where sparse terminal rewards must assign credit to decisions made many steps earlier~\citep{zhou2026human}.
An alternative is to freeze the VLA and train a lightweight MLP specialist only for each precision-critical phase~\citep{xu2026rltoken}.
This avoids RL updates to the large action expert and, with phase-local start conditions and success feedback, shortens the credit-assignment horizon.
The learning advantage comes with deployment costs: a phase-specific design adds controllers as more critical skills are introduced, and handoffs between the VLA and specialists can introduce action discontinuities.
The specialist's own execution can also be jittery, making it unstable to deploy~\citep{yu2026actionresidualsrealworldrobot}.

Local specialists thus simplify RL training but complicate deployment.
This raises a question: \emph{Can this local RL expertise instead repair the VLA itself, without retaining auxiliary controllers at deployment?}
A natural approach is to distill the specialist's executed actions into the VLA.
Prior work shows that RL-generated demonstrations can outperform human data~\citep{xu2024rldg,yang2025beyond,xiao2025pld}.
Yet successful trajectories alone leave two problems unresolved.
First, their action labels record one realization of the specialist's behavior, including stochastic fluctuations and switching artifacts.
Direct imitation treats these variations as targets alongside useful corrections, and can reproduce jitter even after the specialist is removed (Section~\ref{sec:motivation}).
The recorded actions leave unused the expert's ability to propose alternatives and its critic's ability to evaluate them.
Second, the VLA induces a different state distribution from specialist-assisted execution.
Under purely offline distillation, small deviations can therefore lead to states without appropriate supervision and compound into task failure \citep{ross2011dagger}.
Effective transfer requires constructing suitable expert action targets at the states the learner actually encounters.

To address these challenges, we introduce \method{}, a value-guided on-policy distillation framework that transfers local RL expertise into a single VLA.
At each critical-phase observation, \method{} re-queries the relevant specialist for multiple action candidates and uses group-normalized critic scores to construct a value-refined expert action distribution.
A weighted flow-matching objective trains the VLA on these alternatives, rather than on fixed executed-action labels.
The same supervision operator supports both stages of learning: an offline warm start uses observations from specialist-assisted rollouts to establish the critical skills, and on-policy repair applies the operator to observations collected by the VLA itself.
During this second stage, specialists supply action targets without taking control, so supervision follows errors arising under the learner's own execution.
Imitation of replayed VLA actions supports retention of noncritical behavior, using the student's stored predictions during on-policy repair.
Only the repaired VLA is deployed, with no auxiliary specialist or controller switching.

The contributions are:
\begin{itemize}
  \setlength{\itemsep}{2pt}
  \setlength{\parsep}{0pt}
  \item We empirically study specialist deployment and trajectory distillation in long-horizon manipulation. Our analysis shows that successful critical-phase takeover can still produce jittery execution, and identifies inherited action fluctuations and teacher--learner state-distribution mismatch as limitations of direct trajectory distillation.
  \item We introduce \method{}, a unified framework that absorbs local RL expertise into the VLA's action expert. Critic-weighted expert candidates define a value-refined supervision distribution, with the same flow-matching objective supporting offline initialization and on-policy repair on learner-visited states, without specialist takeover during online collection.
  \item Across six real-robot tasks with 30 evaluation trials each, \method{} achieves 100\% success on all six tasks. It deploys a single VLA without auxiliary controllers and reduces mean end-effector jerk on successful trajectories by 29.8--67.5\% relative to RL specialist takeover.
\end{itemize}

\section{Related Work}
\label{sec:related}

\paragraph{Real-world RL and VLA post-training.}
On-policy RL approaches~\citep{jiang2026wovrworldmodelsreliable,chen2025pirl,liu2026can}, improve VLA performance beyond supervised fine-tuning (SFT).
However, these methods are largely evaluated in simulation, because parallel rollouts are not available in real-world settings.
To improve sample efficiency, RLPD~\citep{ball2023rlpd} reuses offline data during online learning, while HIL-SERL~\citep{luo2025hilserl} combines demonstrations and human corrections for real-robot training.
ConRFT~\citep{chen2025conrft} extends this direction to VLAs. 
Direct value optimization of expressive action generators can nevertheless destabilize training.
Residual~\citep{yuan2025decorator} and action-editing methods~\citep{dong2026expoftsampleefficientreinforcementlearning}, ease this optimization burden by training lightweight correction policies.
However, a smaller controller does not by itself shorten the temporal credit-assignment horizon.
RL Token (RLT)~\citep{xu2026rltoken} concentrates online learning on precision-critical phases, where a lightweight MLP actor takes over from the base VLA, localizing credit assignment to short task segments.
This design retains auxiliary controllers and policy handoffs at deployment.
These limitations motivate transferring local RL expertise into a single VLA that executes the entire task without auxiliary controllers.

\paragraph{Distilling Knowledge from data and policy priors.}
Supervised fine-tuning (SFT) on human teleoperation data is a common way to transfer task knowledge into pretrained robot policies~\citep{black2024pi0visionlanguageactionflowmodel}.
HG-DAgger~\citep{kelly2019hgdagger} extends demonstration learning by allowing humans to take over during policy execution and aggregating their corrective actions for subsequent policy updates.
Learned policies provide another source of supervision beyond human demonstrations.
RLDG~\citep{xu2024rldg} trains task-specific RL experts and uses their rollouts to fine-tune generalist policies.
Beyond Human Demonstrations~\citep{yang2025beyond} uses diffusion-based RL to generate smooth, low-variance trajectories for VLA training.
PLD~\citep{xiao2025pld} learns residual specialists that probe generalist failure regions, collects recovery trajectories through hybrid rollouts aligned with the generalist's state distribution, and distills the curated data into the generalist with SFT.
In these methods, knowledge transfer relies on action labels recorded during data collection.
\method{} instead queries the specialist actor and critic to construct supervision at each critical-phase observation: it resamples multiple action chunks and weights them by their relative predicted values to form a value-refined target distribution.
The same supervision rule supports offline initialization and subsequent on-policy distillation on learner-visited states, where specialists provide action proposals and value-based preferences without taking control.

\Needspace{22\baselineskip}
\section{Analysis of Specialist Deployment and Distillation}
\label{sec:motivation}

\begin{wrapfigure}{r}{0.45\textwidth}
  \centering
  \includegraphics[width=\linewidth]{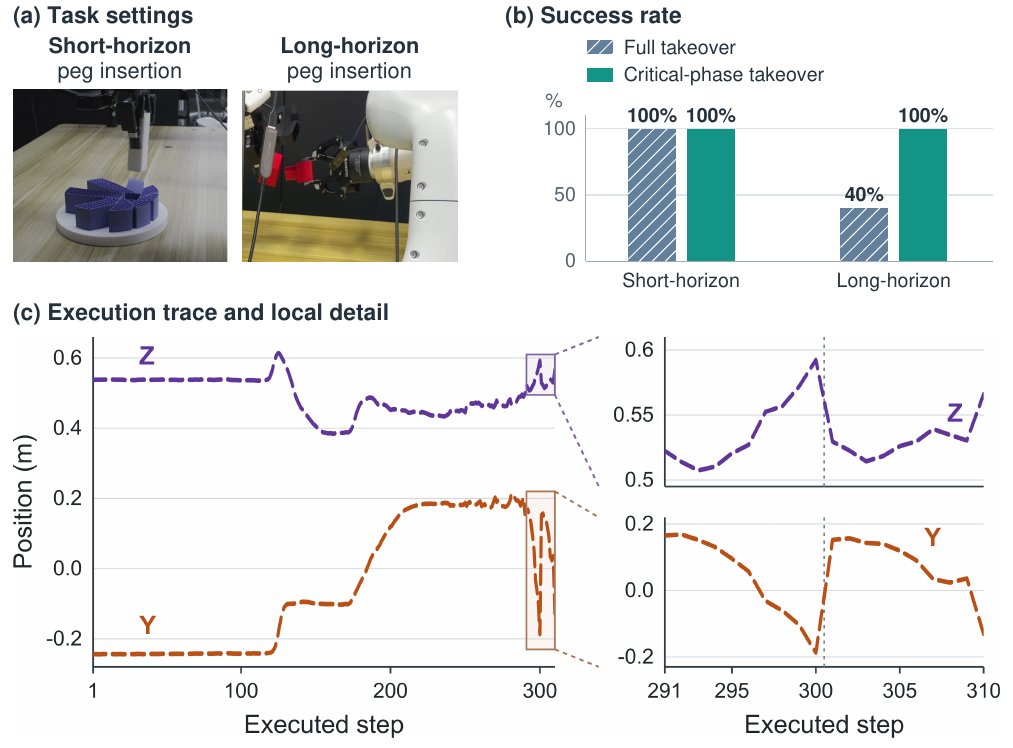}
  \caption{\textbf{Specialist deployment across task horizons.}
  (a) Peg-insertion settings.
  (b) Success rates for full and critical-phase takeover.
  (c) Position traces with local magnification.}
  \label{fig:motivation_scope}
\end{wrapfigure}

\paragraph{Can a lightweight MLP serve as a deployed specialist?}
We examine this question in two peg-insertion settings: short-horizon single-arm insertion and a longer-horizon task culminating in peg assembly.
We use a lightweight MLP specialist trained with RL Token (RLT), and compare two control scopes: \emph{full-task takeover}, where the specialist controls the entire episode, and \emph{critical-phase takeover}, where the base VLA executes the preceding stages and the specialist controls only the final assembly phase.
Our preliminary experiments show comparable success for the two variants in the short-horizon setting.
In the longer-horizon setting, full-task takeover performs poorly, whereas critical-phase takeover achieves high success but exhibits pronounced action jitter (see Figure~\ref{fig:motivation_scope}(c)).
These observations expose a practical tension: restricting the specialist's scope makes the local skill effective, but retaining it as a deployed patch does not yield satisfactory end-to-end control.
They motivate repairing the VLA's own action expert so that the same action generator handles both the preceding stages and precision-critical behavior.

\Needspace{18\baselineskip}
\begingroup
\setlength{\intextsep}{3pt}
\setlength{\abovecaptionskip}{3pt}
\begin{wrapfigure}{r}{0.4\textwidth}
  \centering
  \vspace{-3pt}
  \includegraphics[width=\linewidth]{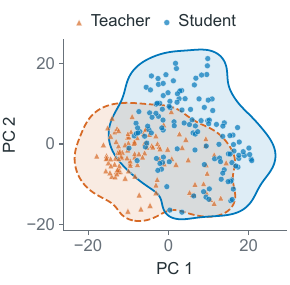}
  \par\vspace{-7pt}
  \caption{\textbf{State coverage after offline distillation.}
  PCA of RL-token observations. Shading marks 95\% KDE regions.}
  \label{fig:motivation_state_coverage}
\end{wrapfigure}

\paragraph{Can the specialist's competence be absorbed into the VLA?}
Direct fine-tuning on 10 specialist trajectories reduces success from 100\% under critical-phase takeover to 60\% (18/30 trials).
The student reproduces the specialist's jitter, copying incidental action fluctuations along with useful corrections.
A successful rollout records one way of completing the task, but does not indicate whether its local action variations were necessary or avoidable.
To examine state coverage, we compare the teacher's and offline-distilled student's RL-token representations using PCA (Figure~\ref{fig:motivation_state_coverage}).
The projected distributions overlap, while student observations also extend beyond the teacher's main region.
Once the student's own actions lead away from the recorded trajectories, it must recover from deviations for which the dataset may contain few corrective examples.
Together, the inherited jitter and coverage mismatch suggest overfitting to a limited set of executions.
Reliable transfer therefore requires refining the action targets and extending supervision to learner-visited states.
\par\WFclear
\endgroup
\vspace{-3pt}

\section{From Local Specialists to Action-Expert Repair}
\label{sec:method}

\method{} repairs the VLA's action expert using value-guided supervision from temporary phase-specific RL specialists.
We first define the repair problem and its supervision requirements, then outline the framework before detailing supervision construction and the two distillation stages.
The same supervision mechanism supports offline initialization and on-policy repair; the two stages differ in where observations are collected.

\subsection{Problem Setup and Supervision Requirements}
\label{sec:problem}

Motivated by Section~\ref{sec:motivation}, we study how to repair a pretrained VLA's action expert using temporary phase-specific RL specialists.
We consider a family of long-horizon manipulation tasks represented as partially observed Markov decision processes.
At time $k$, the policy receives observation $o_k$ (images and proprioception), a language instruction $\ell$, and predicts an action chunk $a_k \in \mathcal{A}^{H}$ of horizon $H$.
The student $\pi_\theta(a\mid o,\ell)$ is the pretrained VLA to be repaired.
A task contains $C\geq1$ precision-critical phases.
Let $g(o,\ell)\in\{0,\ldots,C\}$ denote the experimenter-supplied training-time phase annotation, where $0$ indicates a noncritical observation.
Each critical phase $c\in\{1,\ldots,C\}$ has a lightweight MLP specialist $\pi_{E,c}(a\mid o,\ell)$ and a corresponding critic $Q_{E,c}(o,\ell,a)$.
For an observation with $g(o,\ell)=c>0$, this pair supplies the specialist supervision.

\paragraph{Repair objective.}
We seek a single deployed VLA that completes the entire task with high success, efficient execution, and smooth control under a fixed real-robot interaction budget.
The specialists are available as sources of training data and supervision, but are absent at deployment.
Repair should improve behavior across the critical phases while retaining the VLA's competence in the remaining stages.

\paragraph{Two support gaps.}
Let $\rho_E$ denote the observation distribution in successful specialist or hybrid rollouts and $d^{\pi_\theta}$ the distribution induced when the student acts alone.
Offline trajectory distillation minimizes imitation error under $\rho_E$, but deployment error accumulates under $d^{\pi_\theta}$.
We call this mismatch the \emph{state-support gap}.
At a fixed state, the student may also fail to propose a useful recovery action, and a single executed expert action does not reveal the set of high-value alternatives.
We call this the \emph{action-support gap}.

\paragraph{Supervision requirements.}
These gaps call for supervision that covers the states encountered by the VLA when acting independently and provides useful expert action alternatives at those states.
The supervision should distinguish actions by their expected contribution to task success, rather than treat every recorded or sampled expert action as equally desirable.
The resulting problem is to construct an expert-informed action supervision distribution on learner-induced states that enables the VLA itself to acquire the critical skill and correct its execution errors.

\subsection{Framework Overview}
\label{sec:framework}
\label{sec:rlt_training}

\begin{figure}[!t]
  \centering
  \includegraphics[width=\linewidth]{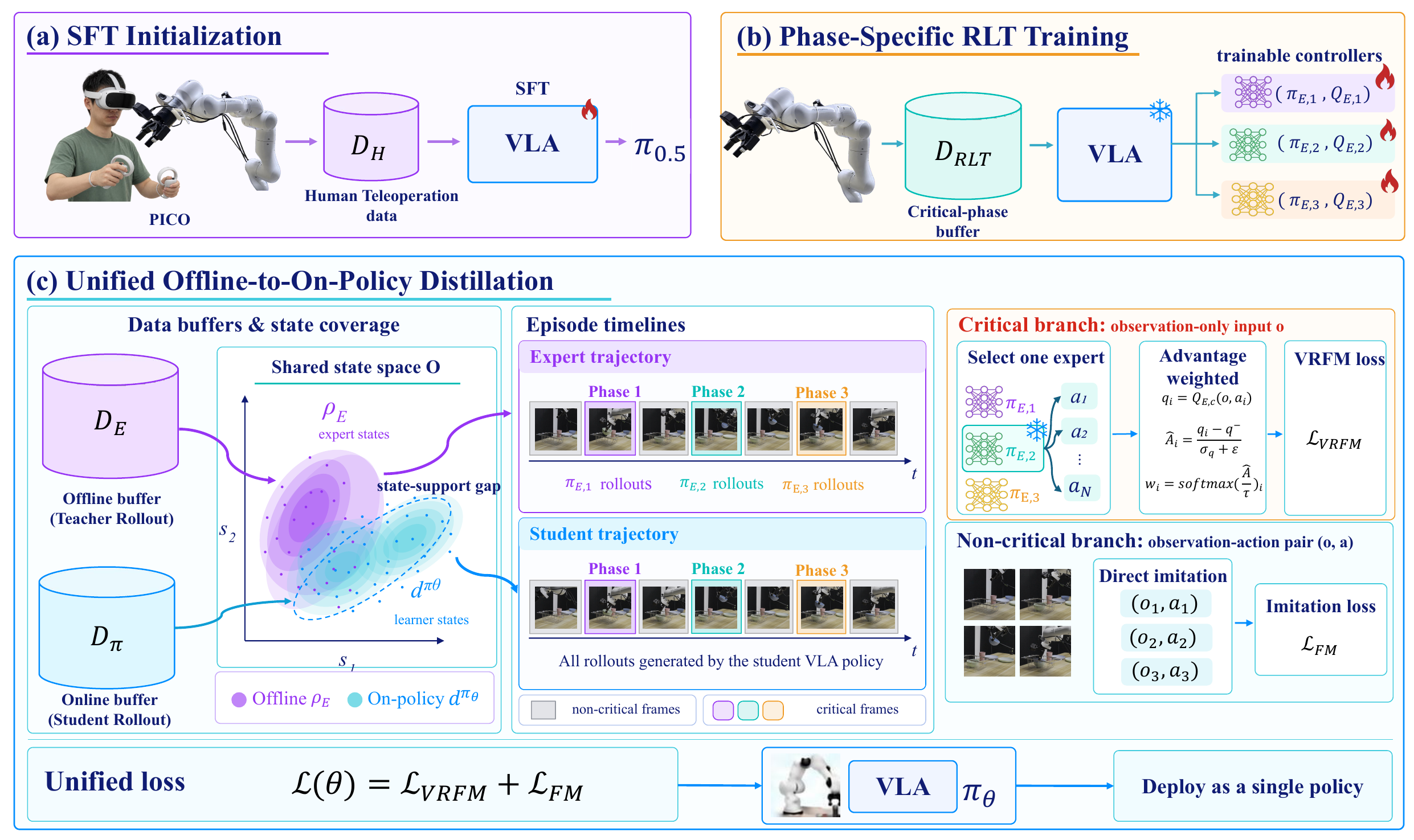}\par
  \vspace{-0.8em}
  \caption{\textbf{Training framework of \method{}.} (a) Human demonstrations initialize the base VLA $\pi_{0.5}$. (b) With this VLA frozen, critical-phase RL trains multiple specialist actors and their critics. (c) Offline observations from $\mathcal{D}_E$ and on-policy observations from $\mathcal{D}_\pi$ feed a shared value-refined supervision mechanism. For each critical observation, the selected phase specialist generates the action candidates and its critic determines their relative weights. The VLA learns from these targets in critical phases and replayed VLA actions in noncritical phases. Only the VLA is deployed.}
  \label{fig:framework}
  \vspace{-1.0em}
\end{figure}

Figure~\ref{fig:framework} organizes \method{} into SFT initialization, phase-specific specialist training, and unified offline-to-on-policy distillation.
In panel (a), human demonstrations $\mathcal{D}_H$ are used to fine-tune a pretrained VLA, yielding the base policy $\pi_{0.5}$ that initializes the student $\pi_\theta$ and supports specialist training.

Panel (b) prepares the local experts using RL Token (RLT) \citep{xu2026rltoken}.
With $\pi_{0.5}$ frozen, we train the lightweight actor $\pi_{E,c}$ and critic $Q_{E,c}$ for each critical phase through robot interaction with human interventions.
The base policy handles noncritical stages, while each specialist learns its local skill.
The trained RL actors are subsequently held fixed, and specialist-assisted rollouts provide observations and phase annotations for the offline buffer $\mathcal{D}_E$.
Specialist-training details are deferred to Appendix~\ref{app:specialist_details}.

Panel (c) connects both observation sources to the same value-refined supervision mechanism.
At a critical observation, the phase annotation selects one specialist, which proposes multiple action candidates; its corresponding critic determines their relative weights in the distillation target.
The VLA's action expert learns from these targets in critical phases and imitates stored VLA actions in noncritical phases.
Offline distillation first draws observations from $\mathcal{D}_E$ to establish the critical skills, generating expert actions anew instead of treating executed actions as fixed labels.

After this warm start, student-only rollouts in panel (c) supply observations to $\mathcal{D}_\pi$.
The specialists provide action candidates and value estimates at these learner-induced states without taking control.
Both stages use the same supervision rule and training objective.
Only the repaired VLA is retained at deployment.

\subsection{Value-Guided Offline-to-On-Policy Distillation}
\label{sec:value_refinement}

\paragraph{Value-refined expert targets.}
For each critical observation, we write $\pi_E$ and $Q_E$ for the specialist and critic selected by $g(o,\ell)$, suppressing the phase subscript in the following equations.
To meet the requirements in Section~\ref{sec:problem}, we construct a value-tilted expert distribution
\begin{equation}
  p_E^Q(a\mid o,\ell)
  \propto
  \pi_E(a\mid o,\ell)
  \exp\!\left(\frac{A_E(o,\ell,a)}{\tau}\right),
  \label{eq:target_measure}
\end{equation}
where $A_E$ is a within-state relative value and $\tau$ controls how sharply the critic emphasizes high-value expert actions.
\method{} approximates this distribution with a finite candidate set and trains the VLA by weighted flow matching.

For a critical observation $(o,\ell)$, draw $N$ action chunks
\begin{equation}
  a_i \sim \pi_E(\cdot\mid o,\ell), \qquad
  q_i = Q_E(o,\ell,a_i), \qquad i=1,\dots,N.
\end{equation}
To remove arbitrary per-state shifts and scales in the critic, compute group-relative scores
\begin{equation}
  \widehat A_i = \frac{q_i-\bar q}{\max(\sigma_q,\epsilon)},
  \qquad
  w_i = \frac{\exp(\widehat A_i/\tau)}
  {\sum_{j=1}^{N}\exp(\widehat A_j/\tau)},
  \label{eq:weights}
\end{equation}
where $\bar q$ and $\sigma_q$ are the mean and standard deviation across candidates, and $\epsilon$ stabilizes the denominator.
The weights emphasize high-value alternatives within the selected specialist's candidate set.

\paragraph{Unified distillation objective.}
Let $\mathcal{L}_{\mathrm{FM}}(\theta;o,\ell,a)$ denote the VLA's standard flow-matching loss for target action chunk $a$ \citep{lipman2023flow}.
We combine the candidate losses using their critic-derived weights:
\begin{equation}
  \mathcal{L}_{\mathrm{VRFM}}(\theta;o,\ell)
  =
  \E_{\{a_i\}\sim\pi_E}
  \left[
  \sum_{i=1}^{N} w_i\,\mathcal{L}_{\mathrm{FM}}(\theta;o,\ell,a_i)
  \right].
  \label{eq:vrfm}
\end{equation}
For noncritical observations, $\mathcal{L}_{\mathrm{FM}}$ imitates the stored action $a^{\mathrm{r}}$: recorded VLA actions offline and student predictions online.
For replay sampling distribution $\rho$, the unified objective is
\begin{equation}
  \mathcal{L}(\theta;\rho)
  = \E_{(o,\ell,a^{\mathrm{r}})\sim\rho}
    [\mathbb{I}\{g(o,\ell)>0\}\mathcal{L}_{\mathrm{VRFM}}
    +\mathbb{I}\{g(o,\ell)=0\}\mathcal{L}_{\mathrm{FM}}].
  \label{eq:distillation}
\end{equation}
Noncritical supervision becomes self-distillation on student rollouts.

\paragraph{From offline initialization to on-policy repair.}
After offline training on $\mathcal{D}_E$, student-only rollouts populate $\mathcal{D}_\pi$ and are added to the same replay.
The critics are frozen offline but updated during on-policy repair to mitigate value-estimation errors on student-visited states.
Equation~\ref{eq:distillation} applies throughout with fixed RL actors, extending repair to learner-induced errors without expert takeover and yielding a single VLA for deployment.
Hyperparameters are provided in Appendix~\ref{app:opd_details}.

\section{Main Experiments}
\label{sec:experiments}

Our main experiments evaluate whether \method{} repairs long-horizon task execution while retaining a single VLA at deployment.
We compare task success and motion smoothness across six real-robot tasks: two short-horizon tasks and four long-horizon tasks.

\begin{figure}[!ht]
  \centering
  \input{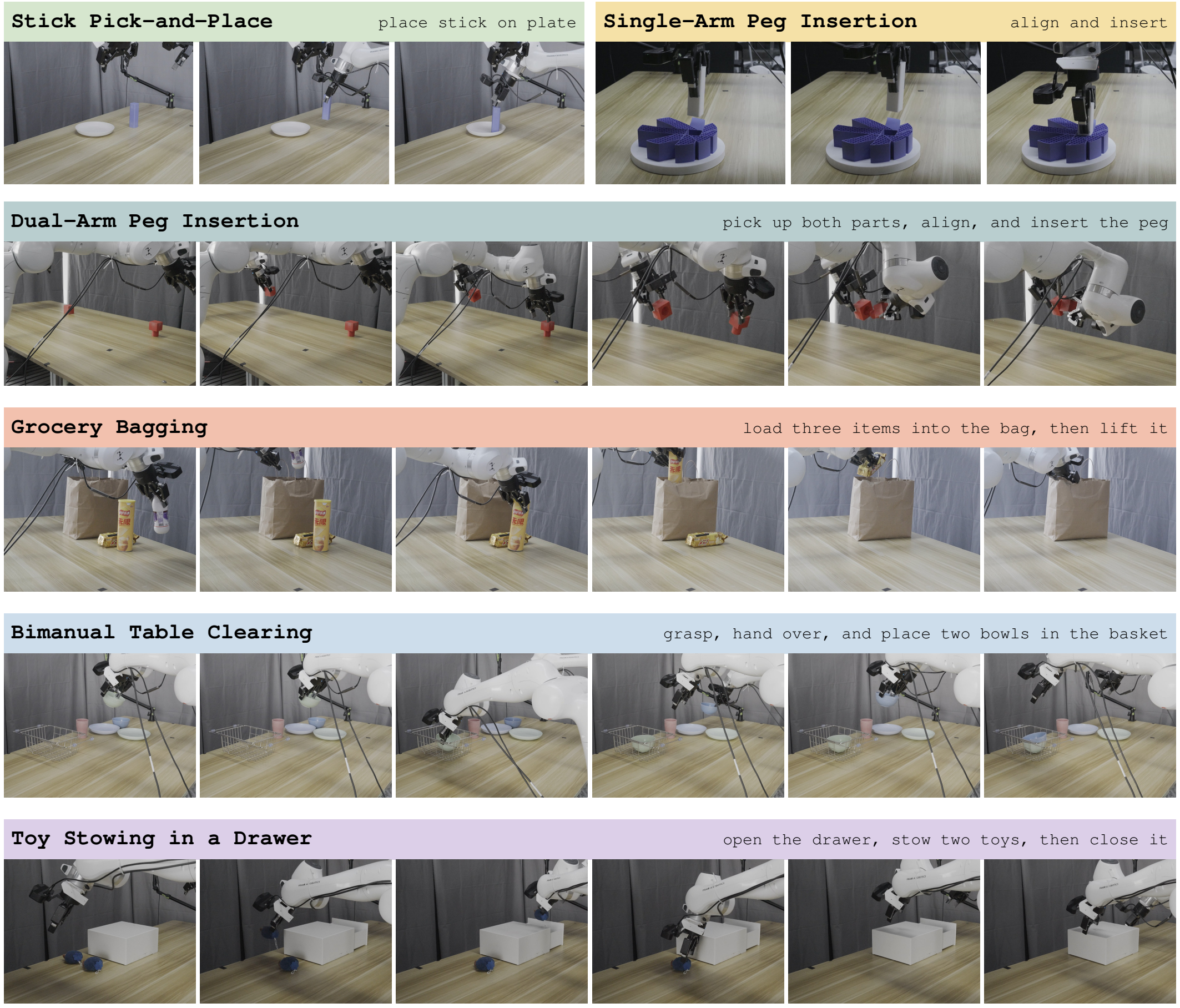}
  \caption{The tasks in our experiments. }
  \label{fig:tasks}
\end{figure}

\paragraph{Tasks and evaluation protocol.}
Figure~\ref{fig:tasks} illustrates the task sequences.
The two short-horizon tasks are \textbf{Stick Pick-and-Place}, which requires grasping a stick and placing it on a plate, and \textbf{Single-Arm Peg Insertion}, which starts with a held peg and requires aligning and inserting it into a fixture.
The four long-horizon tasks involve multiple sequential subgoals:
\textbf{Dual-Arm Peg Insertion} requires picking up the socket and peg with separate arms, aligning the parts, and completing insertion;
\textbf{Grocery Bagging} requires loading a beverage bottle, a chip container, and a biscuit packet into a bag before lifting it by its handles;
\textbf{Bimanual Table Clearing} requires transferring two bowls to a draining basket, with each bowl grasped, handed over between arms, and placed in the basket; and
\textbf{Toy Stowing in a Drawer} requires opening a drawer, placing two toys inside in sequence, and fully closing it.
Each task contains one or more precision-critical phases.
We evaluate each method for 30 trials on each task.
Appendix~\ref{app:task_descriptions} specifies the task configurations, critical phases, success criteria, and representative failure modes.

\paragraph{Compared methods.}
\label{sec:baselines}
We compare \method{} with five baselines:
\textbf{Base SFT}, the VLA fine-tuned on human demonstrations without subsequent RL adaptation;
\textbf{RLT}~\citep{xu2026rltoken}, which retains the base VLA outside critical phases and uses the corresponding RLT specialist within each critical phase;
\textbf{EXPO-FT}~\citep{dong2026expoftsampleefficientreinforcementlearning}, which fine-tunes the VLA via RL using a learned residual policy to refine VLA-generated action chunks, with Q-guided selection;
\textbf{ConRFT}~\citep{chen2025conrft}, which adapts the VLA through offline-to-online reinforcement learning with a consistency-policy objective; and
\textbf{HG-DAgger}~\citep{kelly2019hgdagger}, which aggregates human takeover demonstrations for supervised policy updates.

\paragraph{Training budgets.}
For SFT initialization, all methods use 50 human-teleoperated trajectories per short-horizon task and 100 per long-horizon task (Appendix~\ref{app:human_data_scaling}).
To ensure a fair comparison, we allocate the same adaptation rollout budget of 50 trajectories per task to all methods, excluding human demonstrations.
Each online baseline uses this budget entirely for training, while \method{} distributes it across specialist training and distillation.
Specifically, we first train an RLT teacher, denoted RLT$^{*}$, using 30 online trajectories per task.
We then freeze this checkpoint and collect 10 additional specialist-assisted trajectories for offline distillation, followed by 10 trajectories generated by the student VLA for on-policy distillation.
Thus, including specialist training and data collection for both distillation stages, \method{} uses the same total of 50 adaptation trajectories per task as the baselines.
After distillation, \method{} deploys only the repaired VLA, without specialist takeover or critic-based action selection.

\paragraph{Metrics.}
\textbf{Success rate (SR)} is the percentage of evaluation episodes that satisfy the complete task's success criterion before timeout or a safety abort.
\textbf{Jerk} measures motion smoothness through the time derivative of end-effector acceleration.
For an episode with $T$ control steps, let $\mathbf{x}^{(r)}_t\in\mathbb{R}^3$ be the measured Cartesian position of arm $r$ at time $t\Delta t$, with $R$ active arms.
We use the mean magnitude of the third-order finite difference:
\begin{equation}
  J = \frac{1}{R(T-2)}\sum_{r=1}^{R}\sum_{t=3}^{T}
  \left\|\frac{\mathbf{x}^{(r)}_t-3\mathbf{x}^{(r)}_{t-1}
  +3\mathbf{x}^{(r)}_{t-2}-\mathbf{x}^{(r)}_{t-3}}{\Delta t^3}\right\|_2.
  \label{eq:jerk_metric}
\end{equation}
Jerk is computed over each complete successful episode and reported as the mean $\pm$ standard deviation across successful evaluation trials, in $\mathrm{m\,s^{-3}}$.

\paragraph{Main comparison.}
Table~\ref{tab:main_results} shows that \method{} matches the observed task success of specialist takeover while deploying only a single VLA.
Both \method{} and RLT achieve 30/30 successes on each of the six tasks.
RLT converges early in training, with the 30-trajectory RLT$^{*}$ teacher performing comparably to the 50-trajectory RLT baseline.
We therefore omit a separate entry for RLT$^{*}$ in Table~\ref{tab:main_results}.
The advantage over the other adaptation baselines is most pronounced on long-horizon tasks.
EXPO-FT and ConRFT achieve 80.0--100.0\% success on the two short-horizon tasks, but neither succeeds on Bagging, Table Clearing, or Drawer under the reported training budgets.
HG-DAgger is more competitive on these tasks, yet its aggregate success rate across the four long-horizon tasks is 74.2\%, compared with 100.0\% for \method{}, a gap of 25.8 percentage points.
Alongside this high task success, \method{} reduces mean jerk on successful trajectories by 29.8--67.5\% relative to RLT across all six tasks.
Its jerk values remain close to those of HG-DAgger, which has slightly lower means on four tasks, indicating that the benefit over this imitation baseline lies primarily in higher task success while maintaining a similar level of motion smoothness.

\begin{table}[!ht]
  \vspace{-4pt}
  \centering
  \caption{Main results on six real-robot tasks (30 trials per method per task). SR: success rate (\%, higher is better). Jerk: mean $\pm$ standard deviation over successful trials ($\mathrm{m\,s^{-3}}$, lower is better), with identical sampling and filtering across methods. Dashes denote no successful trials.}
  \label{tab:main_results}
  \scriptsize
  \setlength{\tabcolsep}{1.2pt}
  \begin{tabular*}{\textwidth}{@{\extracolsep{\fill}}l*{6}{rc}@{}}
    \toprule
    & \multicolumn{2}{c}{Stick P\&P} & \multicolumn{2}{c}{Single peg} & \multicolumn{2}{c}{Dual peg} & \multicolumn{2}{c}{Bagging} & \multicolumn{2}{c}{Table clearing} & \multicolumn{2}{c}{Drawer} \\
    \cmidrule(lr){2-3}\cmidrule(lr){4-5}\cmidrule(lr){6-7}\cmidrule(lr){8-9}\cmidrule(lr){10-11}\cmidrule(l){12-13}
    Method & SR $\uparrow$ & Jerk $\downarrow$ & SR $\uparrow$ & Jerk $\downarrow$ & SR $\uparrow$ & Jerk $\downarrow$ & SR $\uparrow$ & Jerk $\downarrow$ & SR $\uparrow$ & Jerk $\downarrow$ & SR $\uparrow$ & Jerk $\downarrow$ \\
    \midrule
    Base SFT ($\pi_{0.5}$) & 30.0 & $4.2 \pm 1.2$ & 33.3 & $3.3 \pm 0.5$ & 0.0 & -- & 13.3 & $4.7 \pm 0.9$ & 20.0 & $4.7 \pm 0.8$ & 0.0 & --  \\
    RLT & 100.0 & $15.7 \pm 2.0$  & 100.0 & $4.7 \pm 0.8$ & 100.0 & $10.3 \pm 0.9$ & 100.0 & $7.2 \pm 0.8$ & 100.0 & $10.4 \pm 2.5$ & 100.0 & $16.7 \pm 5.1$ \\
    EXPO-FT & 80.0 & $13.5 \pm 0.8$ & 93.3 & $4.2 \pm 1.0$ & 33.3 & $9.0 \pm 1.3$ & 0.0 & -- & 0.0 & -- & 0.0 & -- \\
    ConRFT & 93.3 & $12.8 \pm 2.3$ & 100.0 & $3.9 \pm 0.6$ & 20.0 & $7.1 \pm 2.4$ & 0.0 & -- & 0.0 & -- & 0.0 & -- \\
    HG-DAgger & 100.0 & $5.9 \pm 1.2$ & 60.0 & $3.1 \pm 0.6$ & 63.3 & $6.3 \pm 1.1$  & 90.0 & $4.4 \pm 0.8$ & 63.3 & $4.4 \pm 0.6$ & 80.0 & $10.1 \pm 1.8$ \\
    \method{} (ours) & 100.0 & $5.1 \pm 1.0$ & 100.0 & $3.3 \pm 0.7$ & 100.0 & $6.5 \pm 1.6$ & 100.0 & $4.6 \pm 0.9$ & 100.0 & $4.5 \pm 0.7$ & 100.0 & $9.9 \pm 3.9$ \\
    \bottomrule
  \end{tabular*}
  \par\vspace{-4pt}
\end{table}

\Needspace{17\baselineskip}
\begingroup
\setlength{\intextsep}{4pt}
\setlength{\columnsep}{12pt}
\setlength{\abovecaptionskip}{3pt}
\begin{wrapfigure}{r}{0.4\textwidth}
  \centering
  \includegraphics[width=\linewidth]{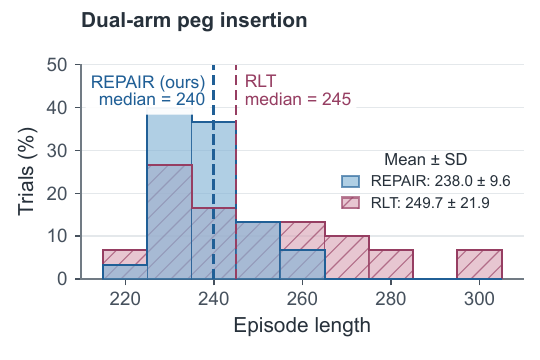}
  \caption{Dual-arm peg insertion episode lengths (30 trials per method; 100\% success).}
  \label{fig:dual_peg_episode_length}
\end{wrapfigure}
We further compare episode lengths on Dual-Arm Peg Insertion to assess whether smoother execution compromises efficiency (Figure~\ref{fig:dual_peg_episode_length}).
Across 30 successful trials per method, \method{} reduces mean episode length by 4.7\%, from $249.7 \pm 21.9$ to $238.0 \pm 9.6$ (mean $\pm$ SD).
Together with 36.9\% lower mean jerk ($6.5$ versus $10.3\,\mathrm{m\,s^{-3}}$), this indicates smoother execution with shorter, less variable episodes.
At a common control timestep, shorter episodes would also imply faster completion.
Fewer alignment corrections during controller handoffs or specialist execution may contribute, although this mechanism remains unverified.
\par\vspace{2\baselineskip}\WFclear
\endgroup

\section{Analysis and Ablations}
\label{sec:results}
\label{sec:ablations}

We ablate two key design choices in \method{}: which actions the VLA learns from and which states it visits during distillation.
The evaluation protocol and metrics follow Section~\ref{sec:experiments}.

\paragraph{RQ1: What supervision should be transferred from a specialist?}
To isolate the effect of supervision from state coverage and online interaction, we first study this question in an offline setting with a fixed expert buffer $\mathcal{D}_E$.
We compare executed specialist actions, one re-queried expert action ($N=1$), eight candidates with uniform weights, and the same candidate sets with critic-based weights, keeping Base SFT initialization, observation sampling, noncritical replay supervision, and optimizer-update budgets fixed.
Executed-action cloning follows the trajectory-supervision principle shared by RLDG~\citep{xu2024rldg} and PLD~\citep{xiao2025pld}: SFT on recorded specialist actions.
It achieves 60.0\% success with the highest mean jerk, $11.1$ (Table~\ref{tab:supervision_ablation}).
Consistent with Section~\ref{sec:motivation}, this suggests that limited-data cloning can overfit abrupt, aggressive RLT corrections alongside useful task behavior.
Re-querying one action improves success to 66.7\% and lowers jerk to $9.2$, while uniform supervision over eight candidates reaches 76.7\% and $7.28$, respectively, suggesting a benefit from broader expert action support.
Critic weighting of the same eight candidates further improves success to 90.0\% and lowers jerk to $6.7$.
Relative to executed-action cloning, this is a gain of 30.0 percentage points in success rate and a 39.6\% reduction in mean jerk on successful trials, supporting the use of multiple expert alternatives with value-based preferences.

To further evaluate value-refined action supervision during on-policy repair, we initialize all variants from the same VLA checkpoint obtained using \method{}'s offline training recipe.
We compare value weighting with online critic updates, value weighting with a frozen critic, and uniform weighting of expert candidates.
The latter two variants initially preserve the offline policy's performance but gradually degrade with continued training.
Online critic updates progressively improve success to 100\% and sustain this level.
These trends suggest that adapting the critic to student-visited states helps maintain effective supervision during on-policy repair.

\begin{table}[!ht]
  \vspace{-4pt}
  \centering
  \begin{minipage}[t]{0.53\textwidth}
    \vspace{0pt}
    \caption{\textbf{Which actions to learn.} Supervision targets on fixed specialist states (Dual Peg; 30 trials). SR in \%; jerk: mean $\pm$ SD over successful trials ($\mathrm{m\,s^{-3}}$).}
    \label{tab:supervision_ablation}
    \centering
    \small
    \setlength{\tabcolsep}{2pt}
    \begin{tabular*}{\linewidth}{@{\extracolsep{\fill}}lccc@{}}
      \toprule
      Action targets & $N$ & SR $\uparrow$ & Jerk $\downarrow$ \\
      \midrule
      Executed actions & -- & 60.0 & $11.1 \pm 1.1$ \\
      Re-queried action & 1 & 66.7 & $9.2 \pm 0.8$ \\
      Uniform candidates & 8 & 76.7 & $7.28 \pm 1.3$ \\
      Value-weighted candidates & 8 & 90.0 & $6.7 \pm 1.1$ \\
      \bottomrule
    \end{tabular*}
  \end{minipage}\hfill
  \begin{minipage}[t]{0.45\textwidth}
    \vspace{0pt}
    \caption{\textbf{Which states to visit.} Specialist versus student rollouts under fixed action supervision (30 trials per task; SR in \%).}
    \label{tab:state_ablation}
    \centering
    \small
    \setlength{\tabcolsep}{1.5pt}
    \begin{tabular*}{\linewidth}{@{\extracolsep{\fill}}lrrrr@{}}
      \toprule
      & \multicolumn{2}{c}{Rollouts} & \multicolumn{2}{c}{SR $\uparrow$} \\
      \cmidrule(lr){2-3}\cmidrule(l){4-5}
      Variant & Specialist & Student & \shortstack{Dual\\peg} & \shortstack{Table\\clearing} \\
      \midrule
      Offline & 10 & 0 & 90.0 & 73.3 \\
      & 20 & 0 & 90.0 & 80.0 \\
      & 30 & 0 & 93.3 & 83.3 \\
      & 50 & 0 & 100.0 & 90.0 \\
      \midrule
      \method{} & 10 & 10 & 100.0 & 100.0 \\
      \bottomrule
    \end{tabular*}
  \end{minipage}
  \par\vspace{-4pt}
\end{table}

\paragraph{RQ2: Where do the gains from on-policy distillation come from?}
To test whether on-policy gains simply reflect more data, we compare larger offline specialist buffers with on-policy distillation on Dual Peg and Table Clearing (Table~\ref{tab:state_ablation}).
All variants use 100 human demonstrations per task for SFT initialization and the same multi-candidate value-weighted supervision.
At 20 total rollouts, offline distillation achieves 90.0\% and 80.0\% success on Dual Peg and Table Clearing, respectively, compared with 100.0\% on both tasks for \method{}'s 10 specialist plus 10 student rollouts.
This contrast supports a benefit from labeling learner-visited states beyond increasing trajectory count.
Expanding offline buffers from 10 to 50 rollouts raises success from 90.0\% to 100.0\% and from 73.3\% to 90.0\%, respectively.
The largest offline buffer thus matches on-policy performance on Dual Peg but remains 10 percentage points lower on Table Clearing, using 40 extra specialist rollouts versus 10 student rollouts.
The pattern is consistent with a coverage-based explanation: larger specialist buffers may cover more recovery states, whereas on-policy distillation directly labels deviations produced as the student's execution errors accumulate.

\Needspace{16\baselineskip}
\section{Conclusion}

We presented \method{}, which distills local RL expertise into a single VLA using critic-weighted expert action candidates.
A shared flow-matching objective supports offline initialization and on-policy refinement on learner-visited states, with specialists providing supervision without takeover.
Across six real-robot tasks, \method{} achieves 100\% success on all six tasks, matching the observed success rates of specialist takeover while reducing mean end-effector jerk on successful trajectories by 29.8--67.5\%.
Ablations support multiple expert candidates with critic-based weighting, and data-scaling results favor on-policy supervision at matched rollout counts on the analyzed tasks.
Training still relies on phase annotations, and critic rankings on learner states need direct validation.
These findings support transferring specialists' critical skills into the VLA while removing auxiliary controllers at deployment.

\bibliography{references}
\bibliographystyle{plainnat}

\newpage
\appendix
\section{ Franka Robot Setup}
\label{app:franka_setup}

We conduct real-world experiments on a bimanual platform with two 7-DoF Franka arms, each equipped with a parallel-jaw gripper. The policy receives one third-person RGB view, two wrist-mounted RGB views, the language instruction, and the proprioceptive state of both arms. For each arm, the state consists of the TCP position, a 6D rotation representation, and the gripper state, resulting in a 20-D bimanual state
\[
\bm{s}_t =
[\bm{p}^L_t,\bm{r}^L_{6D,t},g^L_t,
 \bm{p}^R_t,\bm{r}^R_{6D,t},g^R_t].
\]
The policy predicts 20-D bimanual actions with the same per-arm layout. While the robot interface executes absolute TCP waypoints, we train $\pi_{0.5}$ using body-frame SE(3) relative actions,
\[
\Delta \bm{T}^{i}_{t,k}
=
(\bm{T}^{i}_{t})^{-1}\bm{T}^{i}_{t+k},
\qquad i\in\{L,R\},
\]
and recover absolute TCP targets before execution. We use an action horizon of $H=10$, and execute the bimanual policy at $10\,\mathrm{Hz}$.

\section{Implementation Details}
\label{app:implementation_details}

\subsection{REPAIR}
\label{app:opd_details}

We implement \method{} with RLinf~\citep{yu2026rlinf,zang2025rlinf,zang2026rlinfuse}.
During distillation, we update the VLA's action expert while keeping the backbone, RL-token encoder, and specialist actors frozen.
The RLT critics are frozen during offline distillation and updated alongside the VLA during online repair to mitigate Q-value overestimation at previously unseen states.
A frozen copy of $\pi_{0.5}$ supplies the specialists' features and reference actions.
Candidate scores use the minimum of the two target critics, and replay minibatches contain equal numbers of critical and noncritical observations.
Table~\ref{tab:opd_recipe} lists the distillation hyperparameters.

\begin{table}[!ht]
  \centering
  \small
  \caption{Distillation hyperparameters for \method{}. Batch size counts observations before candidate expansion.}
  \label{tab:opd_recipe}
  \begin{tabular}{ll}
    \toprule
    Setting & Value \\
    \midrule
    Specialist candidates per critical observation & 8 \\
    Q-weight temperature $\tau$ & 0.1 \\
    Reference dropout probability & 0.5 \\
    Optimizer & AdamW \\
    Learning rate / schedule & $5\times10^{-6}$ / constant \\
    AdamW $(\beta_1,\beta_2)$ / $\epsilon$ & $(0.9,0.95)$ / $10^{-8}$ \\
    Weight decay & 0.01 \\
    Global batch size & 64  \\
    Optimizer steps per learner iteration & 1 \\
    Computation precision & BF16 \\
    \bottomrule
  \end{tabular}
\end{table}

\subsection{RLT}
\label{app:algorithm}
\label{app:specialist_details}

RLT trains phase-specific actors and critics while keeping the base VLA $\pi_{0.5}$ and its RL-token encoder frozen.

\paragraph{Architecture.}
Each critical phase $c\in\{1,\ldots,C\}$ has an actor $\pi_{E,c}$ taking $[\operatorname{vec}(\bar a_t),z_t,p_t]$ and twin critics taking $[z_t,p_t,\operatorname{vec}(a_t)]$.
Both use three 512-unit hidden layers with tanh activations; the critics additionally use layer normalization.
Specialist and reference chunks have horizon $H=10$ and share the replay action coordinates.
The actor samples
\begin{equation}
  \operatorname{vec}(a_t)
  =\tanh\!\left(\mu_c(\bar a_t,z_t,p_t)+\sigma\epsilon\right),
  \qquad \epsilon\sim\mathcal{N}(0,I).
  \label{eq:rlt_actor_sample}
\end{equation}

\paragraph{Phase-local replay.}
The phase annotation activates the frozen VLA for $g=0$ and specialist $c$ for $g=c>0$, with human overrides allowed.
Replay stores executed actions and intervention indicators from successful and failed critical-phase segments.
Each segment terminates on local success or failure, with reward one for local success and zero otherwise.
Later phase outcomes do not alter earlier rewards, and noncritical segments are excluded.

\paragraph{Critic update.}
For $s_t=(z_t,p_t)$, phase-termination indicator $d_t$, and next action $a'\sim\pi_{E,c}(\cdot\mid s_{t+H},\bar a_{t+H})$, the chunk TD target is
\begin{equation}
  y_t=\sum_{j=0}^{H-1}\gamma^j r_{t+j}
  +(1-d_t)\gamma^H\min_{k\in\{1,2\}}
  Q^-_{c,k}(s_{t+H},a'),
  \qquad \gamma=0.96,
  \label{eq:rlt_td_target}
\end{equation}
where $Q^-_{c,k}$ denotes a target critic.
Both critics regress to $y_t$ on executed replay chunks, and target parameters use an exponential moving average with coefficient $\eta=0.005$ after each critic step.
Neither the TD target nor the actor objective includes an entropy term.

\paragraph{Actor update and intervention supervision.}
The actor maximizes the first critic with a behavior-cloning (BC) penalty.
The target $b_{t,j}$ is the executed action at steps containing human intervention and the frozen VLA reference action otherwise.
With validity mask $m_{t,j}$ and actor sample $\widetilde a_t$ from Equation~\ref{eq:rlt_actor_sample}, the loss is
\begin{equation}
  \mathcal{L}_{\mathrm{actor}}
  =-\lambda_Q\,\mathbb{E}\!\left[Q_{c,1}(s_t,\widetilde a_t)\right]
  +\lambda_{\mathrm{BC}}\,
  \frac{\sum_{t,j}m_{t,j}\|\widetilde a_{t,j}-b_{t,j}\|_2^2}
       {d_a\sum_{t,j}m_{t,j}},
  \qquad \lambda_Q=\lambda_{\mathrm{BC}}=1.
  \label{eq:rlt_actor_loss}
\end{equation}
Short terminal chunks use repeat-last padding.
Reference dropout zeros the entire actor reference input with probability $0.5$ during actor updates and training rollouts; BC targets and next-action sampling for the critic target retain the full reference.

\begin{table}[ht]
  \centering
  \small
  \caption{RLT training hyperparameters.}
  \label{tab:rlt_recipe}
  \begin{tabular}{ll}
    \toprule
    Setting & Value \\
    \midrule
    RL-token  & 2048 \\
    proprioceptive dimensions & 20 \\
    Action dimension & 20 \\
    chunk horizon & 10 \\
    Actor / critic hidden widths & $(512,512,512)$ / $(512,512,512)$ \\
    Reference dropout probability & 0.5 \\
    Q / BC objective weights & 1 / 1 \\
    Discount / target-update coefficient & 0.96 / 0.005 \\
    Actor / critic learning rate & $10^{-4}$ / $10^{-4}$ (constant) \\
    Actor / critic gradient-norm clipping & 10 / 10 \\
    Minibatch size & 256 \\
    Critic-to-actor update ratio & $4{:}1$ \\
    \bottomrule
  \end{tabular}
\end{table}

\subsection{Other Baselines}
\label{app:baseline_implementation}

\paragraph{EXPO-FT.}
Following the official implementation~\citep{dong2026expoftsampleefficientreinforcementlearning}, we train a lightweight edit policy over VLA action chunks and select the highest-Q candidate from the base and edited actions.
The VLA is updated with its original supervised objective on replay data, including human corrections.

\paragraph{ConRFT.}
Following the official implementation~\citep{chen2025conrft}, we initialize the consistency policy and critic on offline demonstrations using behavior cloning and calibrated Q-learning.
Online training retains the consistency-based actor objective and uses demonstrations, policy rollouts, and human interventions.

\clearpage
\section{Task Descriptions and Demonstration Budgets}
\label{app:tasks_and_budget}

\begingroup
\raggedbottom
\subsection{Task Descriptions}
\label{app:task_descriptions}

All six tasks use the Franka FR3 platform with Robotiq grippers described in Appendix~\ref{app:franka_setup}.
The descriptions below specify the initial conditions, task sequence, success criteria, precision-critical phases, and representative failure modes.
Task names follow Section~\ref{sec:experiments}.

\newenvironment{appendixtask}{\par\medskip\noindent\begin{minipage}{\linewidth}}{\end{minipage}\par}

\begin{appendixtask}
\paragraph{Single-Arm Peg Insertion.}
A white rectangular peg must be inserted into a matching opening in an irregularly shaped fixture secured to the table.
The peg starts in the gripper, with horizontal and vertical offsets from the opening of approximately $20\,\mathrm{cm}$ or less.
The initial configuration is fixed across episodes.
The robot approaches the opening, aligns the peg, and advances it to full insertion depth.
Success requires the peg to be fully seated at the bottom of the opening within the allotted time.
The critical phase spans initial alignment through completed insertion, covering nearly the entire task.
Precise positional and angular alignment must be maintained throughout: misalignment can prevent entry or obstruct advancement after partial insertion.
Typical failures are failure to enter the opening and incomplete insertion before timeout (Figure~\ref{fig:app_single_peg_failure}).

\begin{center}
  \centering
  \includegraphics[width=0.52\linewidth]{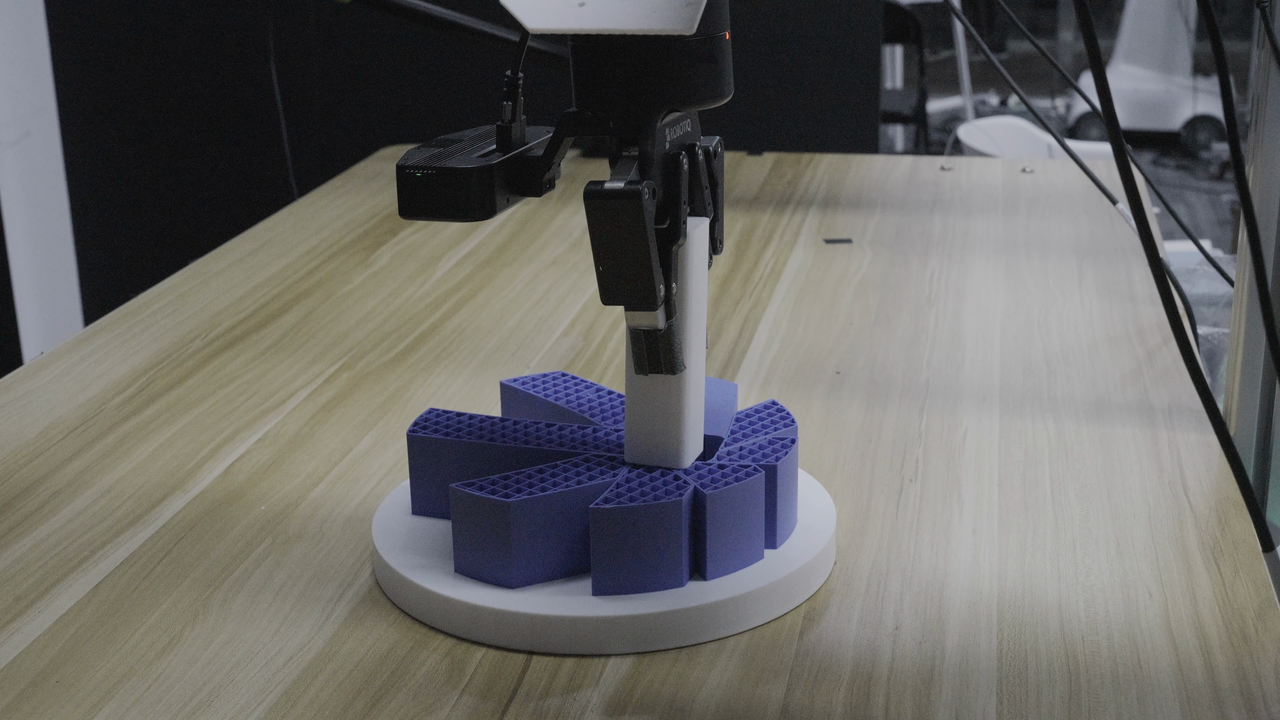}
  \captionof{figure}{Single-Arm Peg Insertion failure. Misalignment prevents the peg from entering the opening and completing insertion within the allotted time.}
  \label{fig:app_single_peg_failure}
\end{center}
\end{appendixtask}

\begin{appendixtask}
\paragraph{Stick Pick-and-Place.}
The robot grasps a purple rectangular stick and places it inside a white circular plate fixed to the table.
The stick's initial position is randomized within the marked semi-annular region around the plate (Figure~\ref{fig:app_stick_setup_failure}, left).
Its location determines the executing arm: the left arm handles the left side and the right arm the right side.
The selected arm grasps and lifts the stick, transports it to the plate, and releases it inside.
Success requires both pickup and placement.
The critical phase is grasp acquisition, from final approach through gripper closure and initial lifting.
This combines spatially conditioned arm selection with accurate gripper--object alignment and a stable grasp across randomized positions.
The predominant failure is an insecure grasp that prevents lifting and subsequent transfer (Figure~\ref{fig:app_stick_setup_failure}, right).

\begin{center}
  \centering
  \includegraphics[height=1.45in]{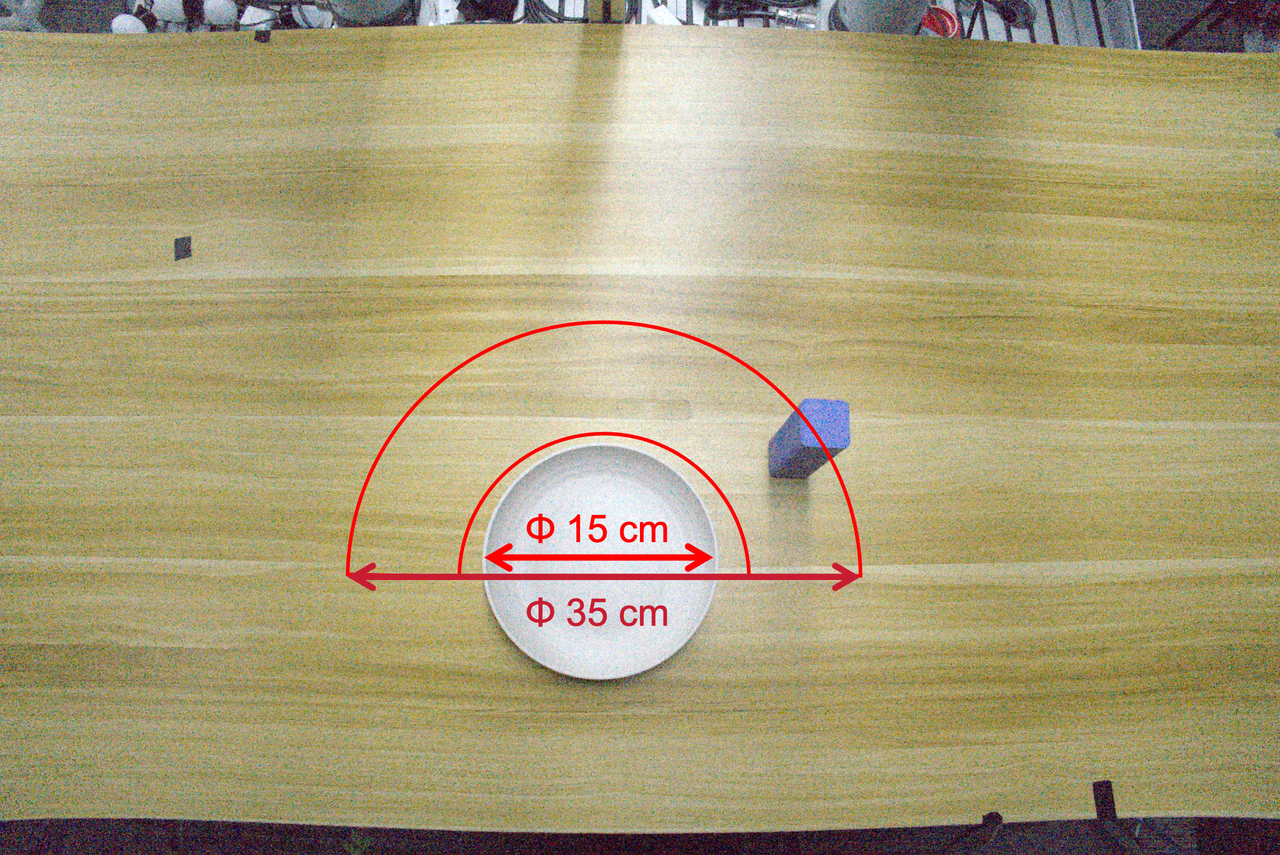}\hfill
  \includegraphics[height=1.45in]{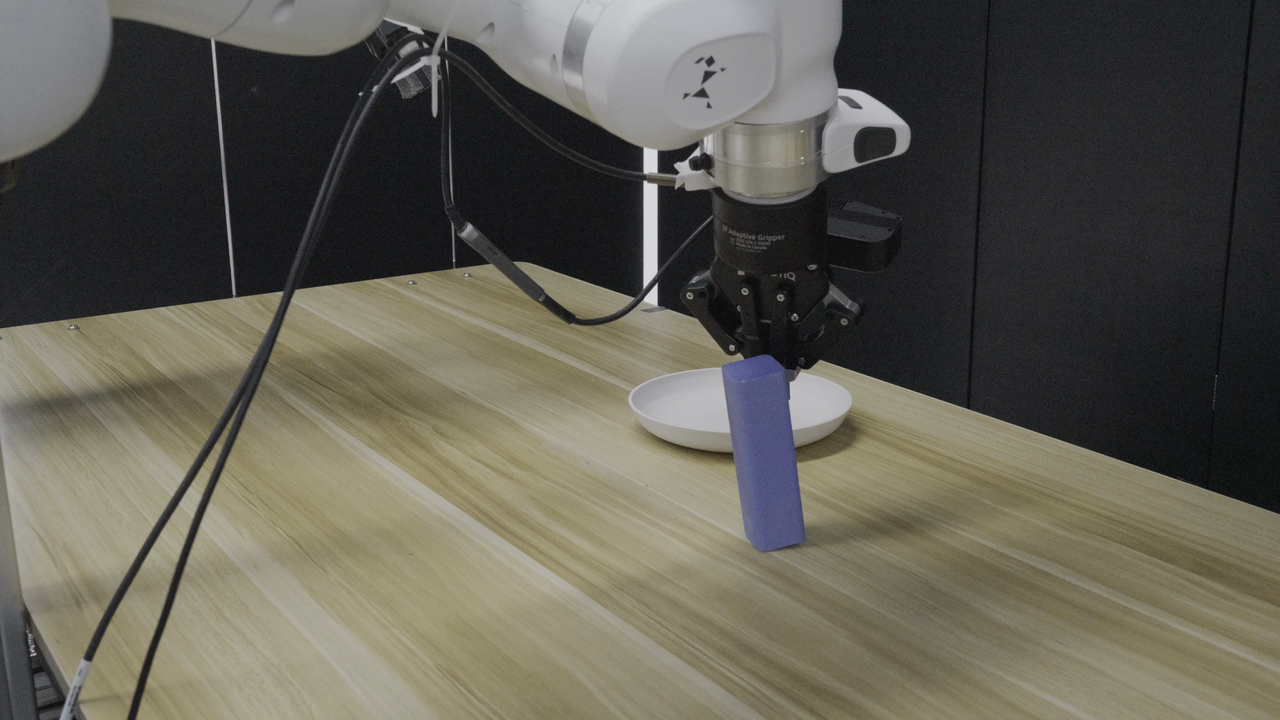}
  \captionof{figure}{Stick Pick-and-Place. Left: the red boundaries mark the object initialization region. Right: the gripper fails to establish a secure grasp. Dimensions and region annotations are retained from the supplied task illustration.}
  \label{fig:app_stick_setup_failure}
\end{center}
\end{appendixtask}

\begin{appendixtask}
\paragraph{Dual-Arm Peg Insertion.}
The robot assembles two complementary components, one with a protruding peg and the other with a matching socket.
The left arm first grasps and lifts the socket, followed by the right arm grasping and lifting the peg.
The robot then brings the components together, aligns them, and performs insertion while holding both.
Success requires completed assembly within the allotted time.
The critical phase comprises relative alignment and insertion after both grasps.
Motion of either arm changes the relative pose, so this phase requires coordinated positional and angular control while maintaining stable grasps.
Failures include repeated unsuccessful insertion attempts until timeout (Figure~\ref{fig:app_dual_peg_failure}) and premature release by the right gripper, causing the peg component to fall before engagement.

\begin{center}
  \centering
  \includegraphics[width=0.52\linewidth]{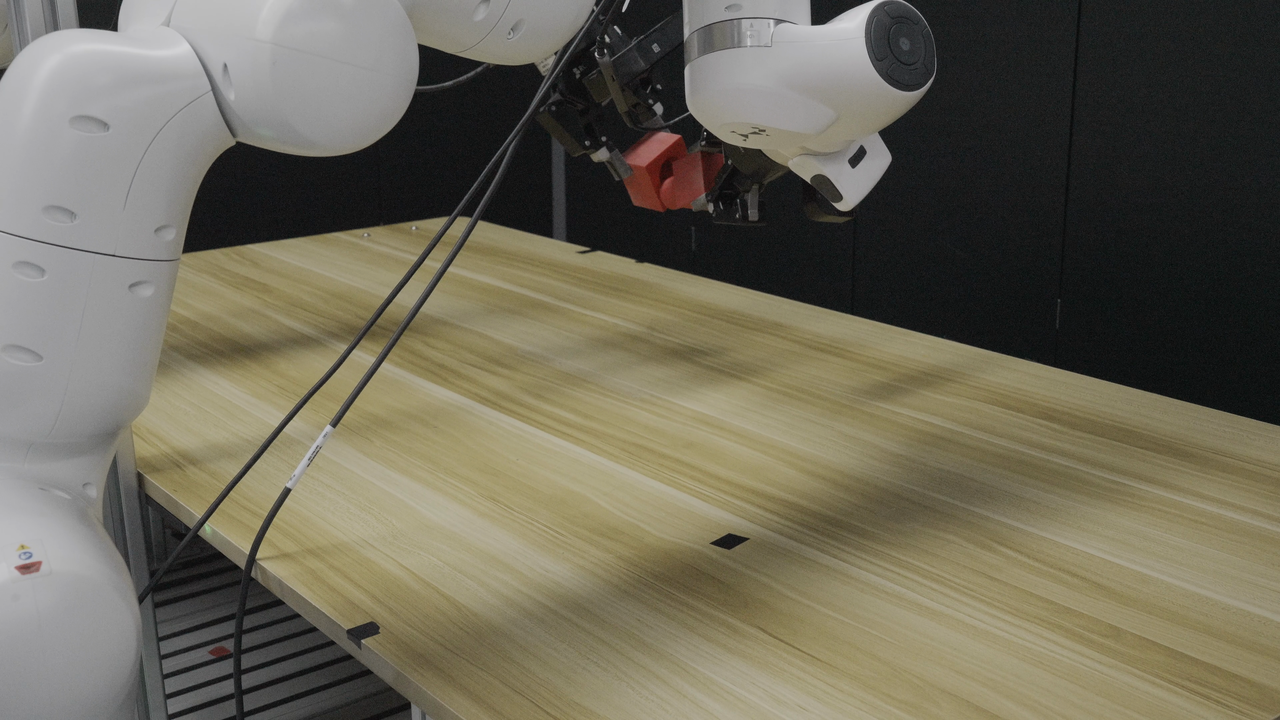}
  \captionof{figure}{Dual-Arm Peg Insertion failure. Misalignment between the held components prevents the peg from entering the socket despite repeated attempts.}
  \label{fig:app_dual_peg_failure}
\end{center}
\end{appendixtask}

\begin{appendixtask}
\paragraph{Toy Stowing in a Drawer.}
Two plush toys start beside a drawer unit secured to the tabletop.
The right arm opens the drawer, the left arm picks up and places the toys inside sequentially, and the right arm closes the drawer.
Success requires both toys to be contained inside and the drawer to be fully closed.
There are two critical phases: placement of the second toy by the left arm and final drawer closure by the right arm.
The second placement must deposit the toy inside the drawer, while closure must complete the drawer's full travel.
Typical failures are depositing a toy outside the drawer and leaving the drawer partially open (Figure~\ref{fig:app_drawer_failures}).

\begin{center}
  \centering
  \includegraphics[width=0.49\linewidth]{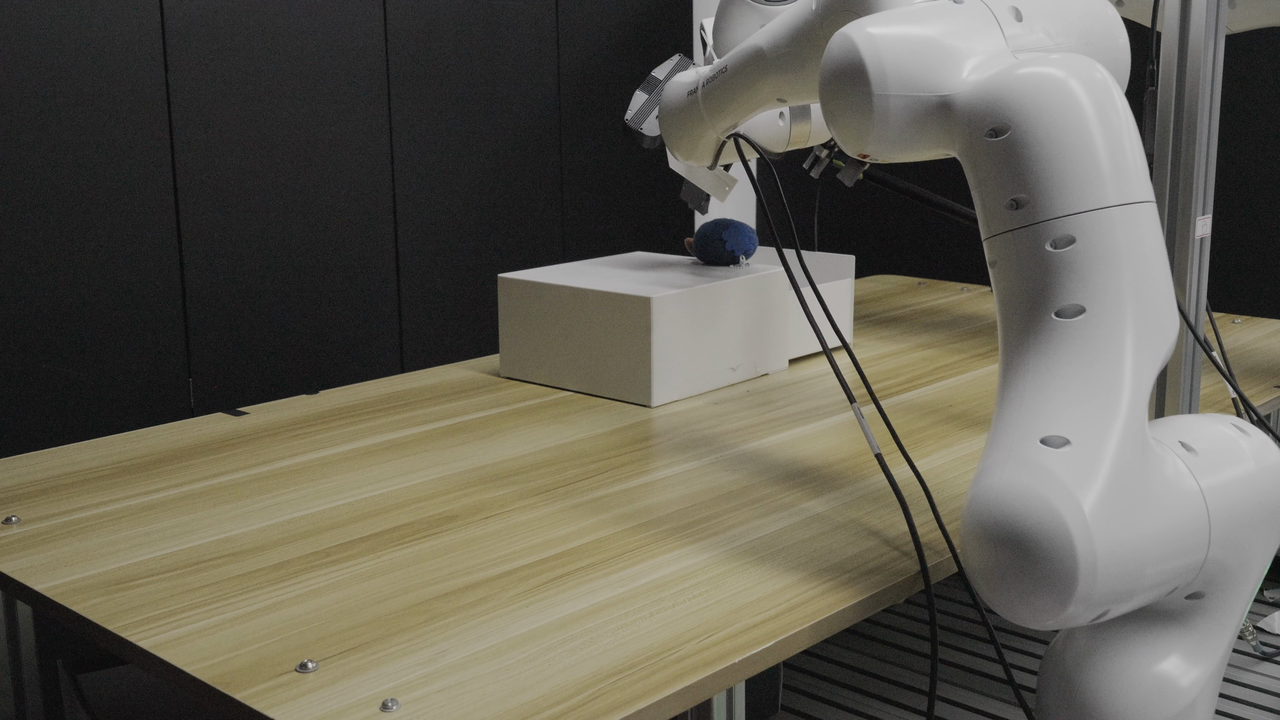}\hfill
  \includegraphics[width=0.49\linewidth]{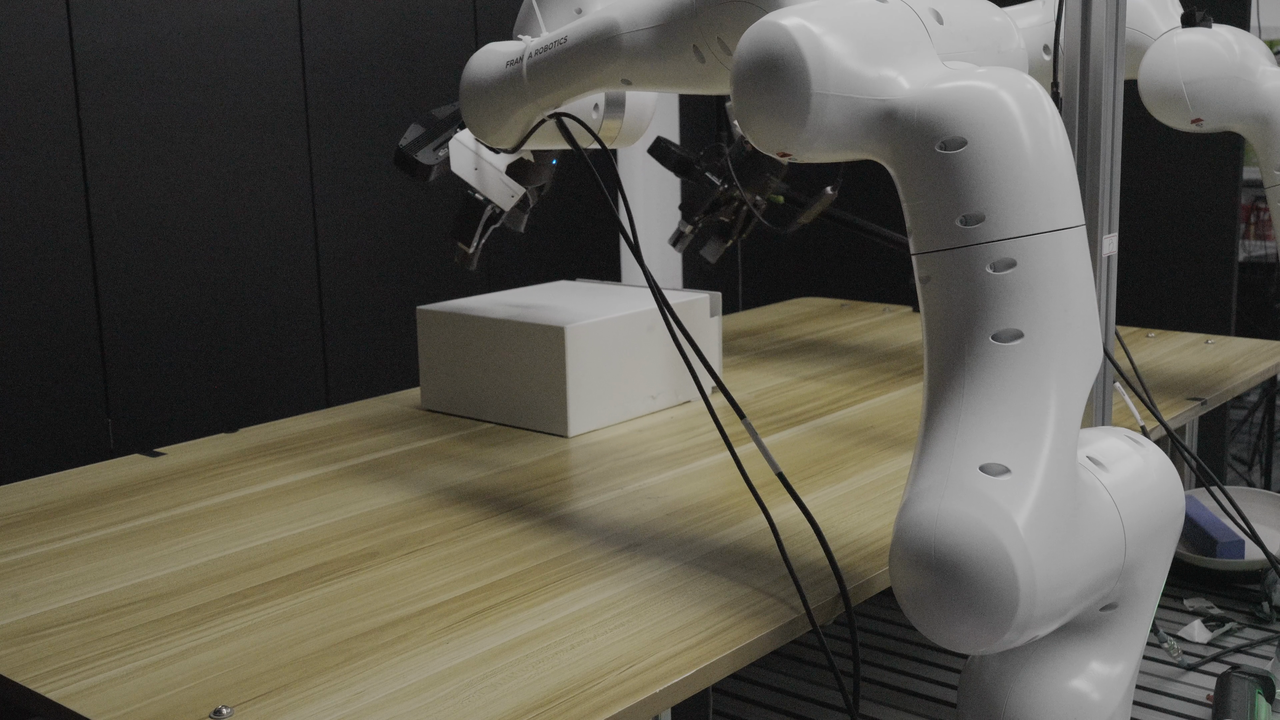}
  \captionof{figure}{Toy Stowing in a Drawer failures. Left: the toy is deposited on top of the drawer housing. Right: the drawer remains partially open after the closing attempt.}
  \label{fig:app_drawer_failures}
\end{center}
\end{appendixtask}

\begin{appendixtask}
\paragraph{Grocery Bagging.}
The robot packs a beverage bottle, a cylindrical chip container, and a biscuit packet into a bag, then lifts it by both handles.
The bag starts at a fixed location; object positions are randomized within the annotated region in Figure~\ref{fig:app_bagging_setup_failures} (top left), with the biscuit packet lying flat.
The left arm raises and holds the left handle while the right arm places the bottle, chip container, and biscuit packet into the bag in sequence.
The right arm then grasps the right handle, and both arms lift the loaded bag.
Success requires all three objects inside the bag and lifting by both handles.
The three critical phases are grasping the chip container, placing it into the bag, and placing the biscuit packet.
The cylindrical container requires precise two-finger grasping, and placements require alignment with a bag opening that may deform after unintended contact.
Failures include an insecure chip-container grasp and inaccurate placement that leaves the container or biscuit packet outside the bag.
During placement, a chip packet may also press against the right rim of the bag instead of passing through the opening, preventing successful insertion.

\begin{center}
  \newcommand{\baggingpanel}[1]{%
    \begin{minipage}[c][0.276\linewidth][c]{0.49\linewidth}
      \centering
      \includegraphics[width=\linewidth,height=0.563\linewidth,keepaspectratio]{#1}%
    \end{minipage}%
  }
  \baggingpanel{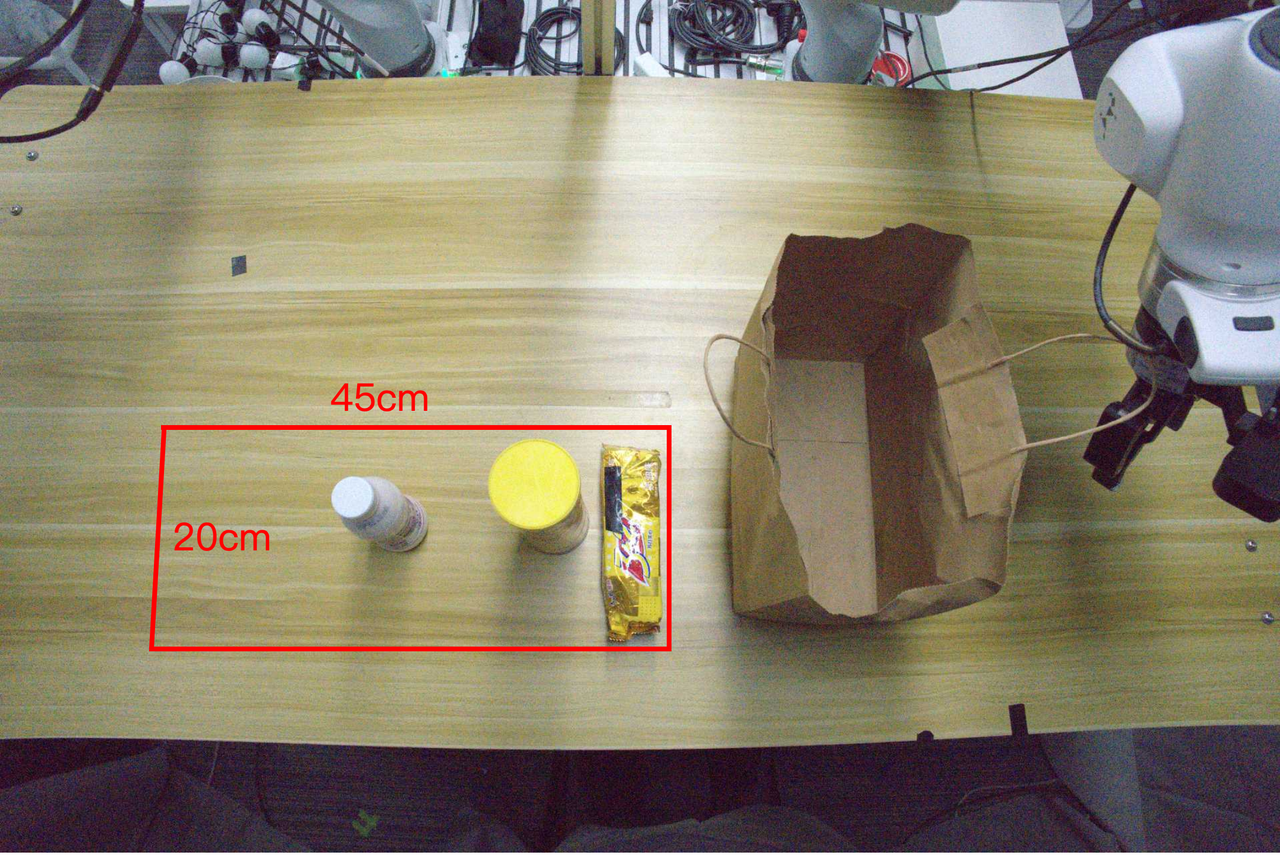}\hfill
  \baggingpanel{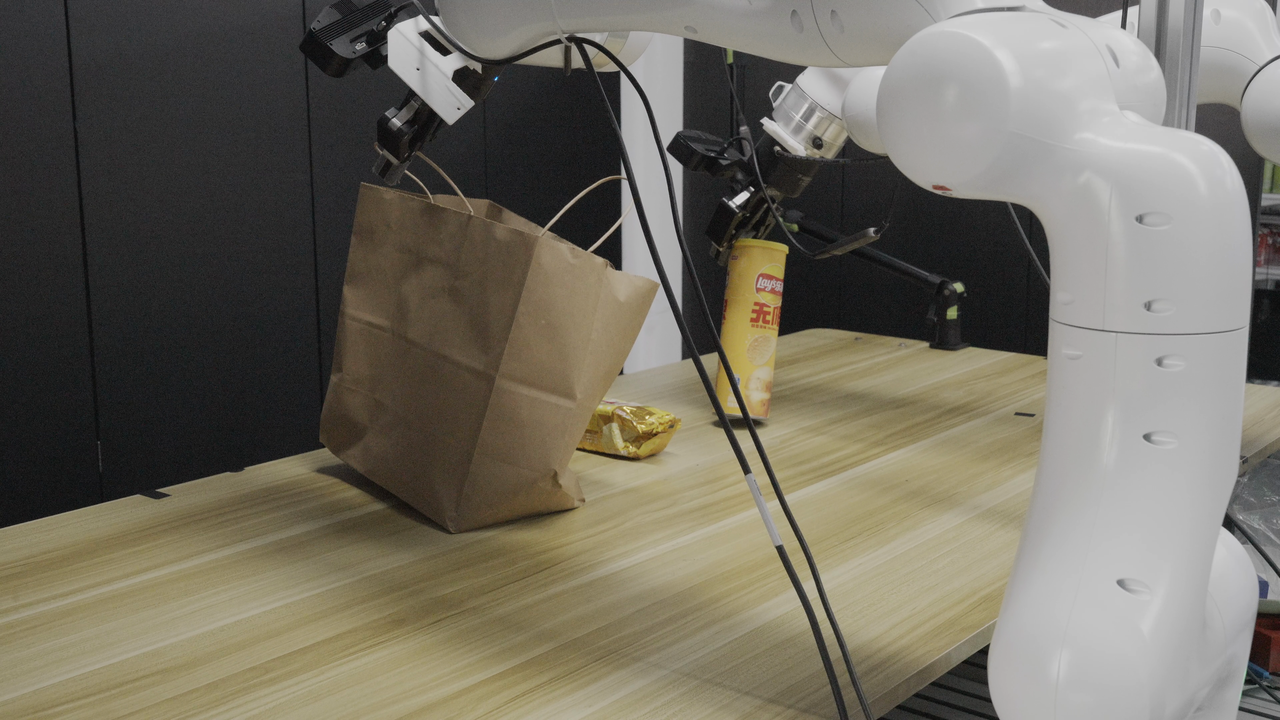}\par\smallskip
  \baggingpanel{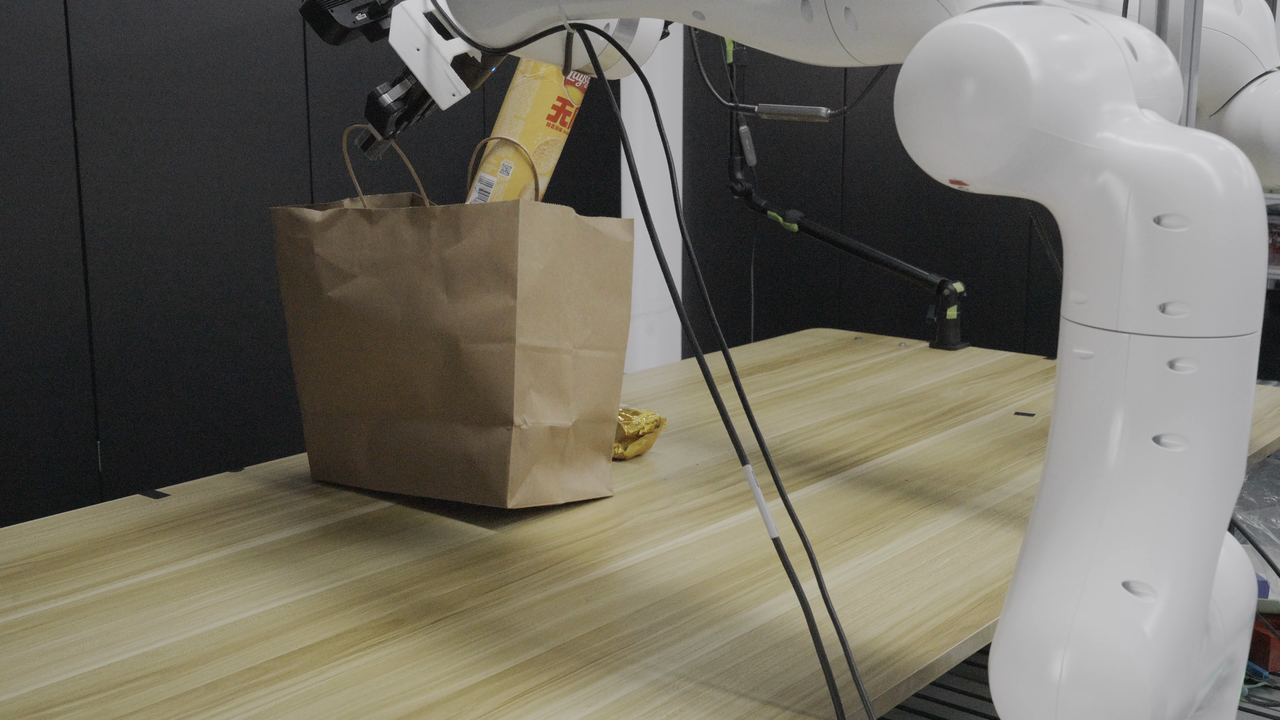}\hfill
  \baggingpanel{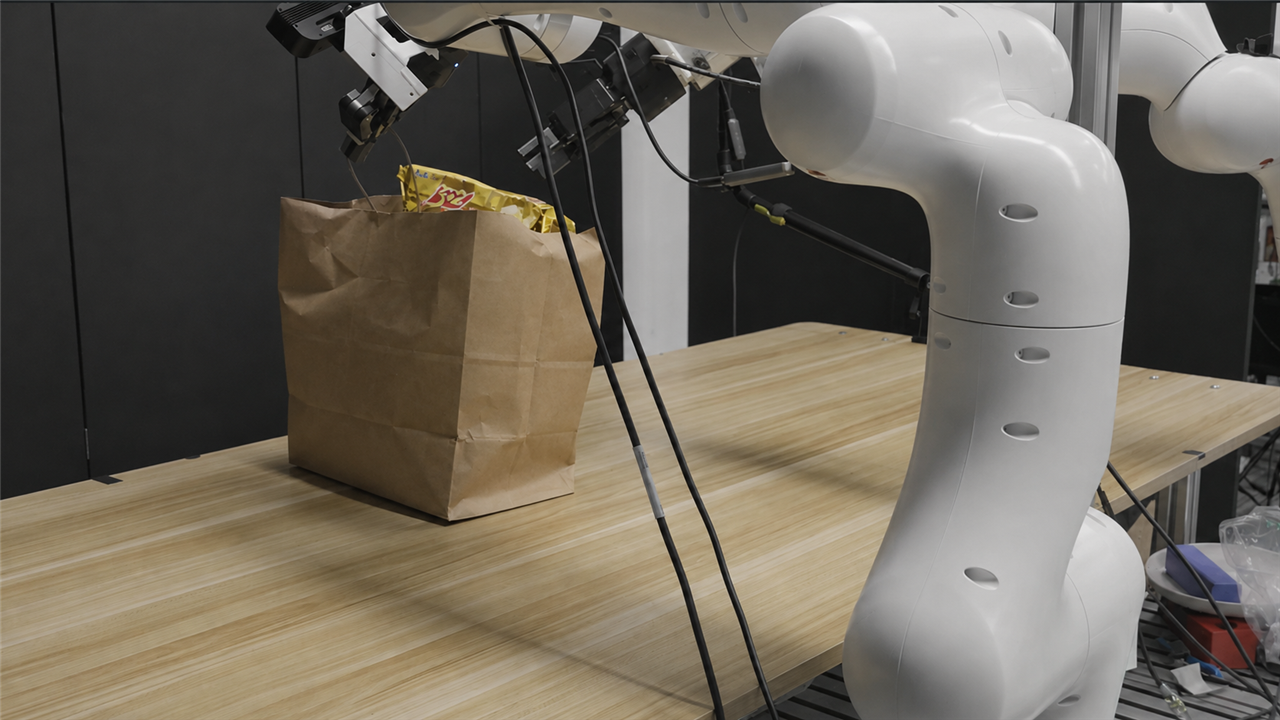}
  \captionof{figure}{Grocery Bagging setup and failures. Top left: the red region defines randomized object initialization. Top right: the right gripper fails to secure the chip container. Bottom left: inaccurate alignment prevents placing the container through the bag opening. Bottom right: a chip packet presses against the right rim of the bag and fails to enter. The bottom-right panel is an AI-enhanced illustration based on the supplied experimental frame.}
  \label{fig:app_bagging_setup_failures}
\end{center}
\end{appendixtask}

\begin{appendixtask}
\paragraph{Bimanual Table Clearing.}
Two bowls must be transferred from the table to a dish-drying basket through sequential handovers.
The setup also includes two plates and a cup.
The bowls start at randomized positions within the marked region in Figure~\ref{fig:app_table_setup_failures} (top); the other items and basket remain fixed.
The right arm picks up the green bowl and hands it to the left arm for placement in the basket.
The sequence is repeated with the blue bowl, which is placed on top of the green bowl.
Success requires both bowls to be stacked inside the basket.
The two critical phases are grasping the blue bowl and transferring it from right to left.
During handover, the left gripper must securely engage the rim before the right gripper releases it.
Failures include missing the blue bowl during pickup and a shallow or missed receiving grasp that lets the bowl fall upon release.

\begin{center}
  \centering
  \includegraphics[width=0.57\linewidth]{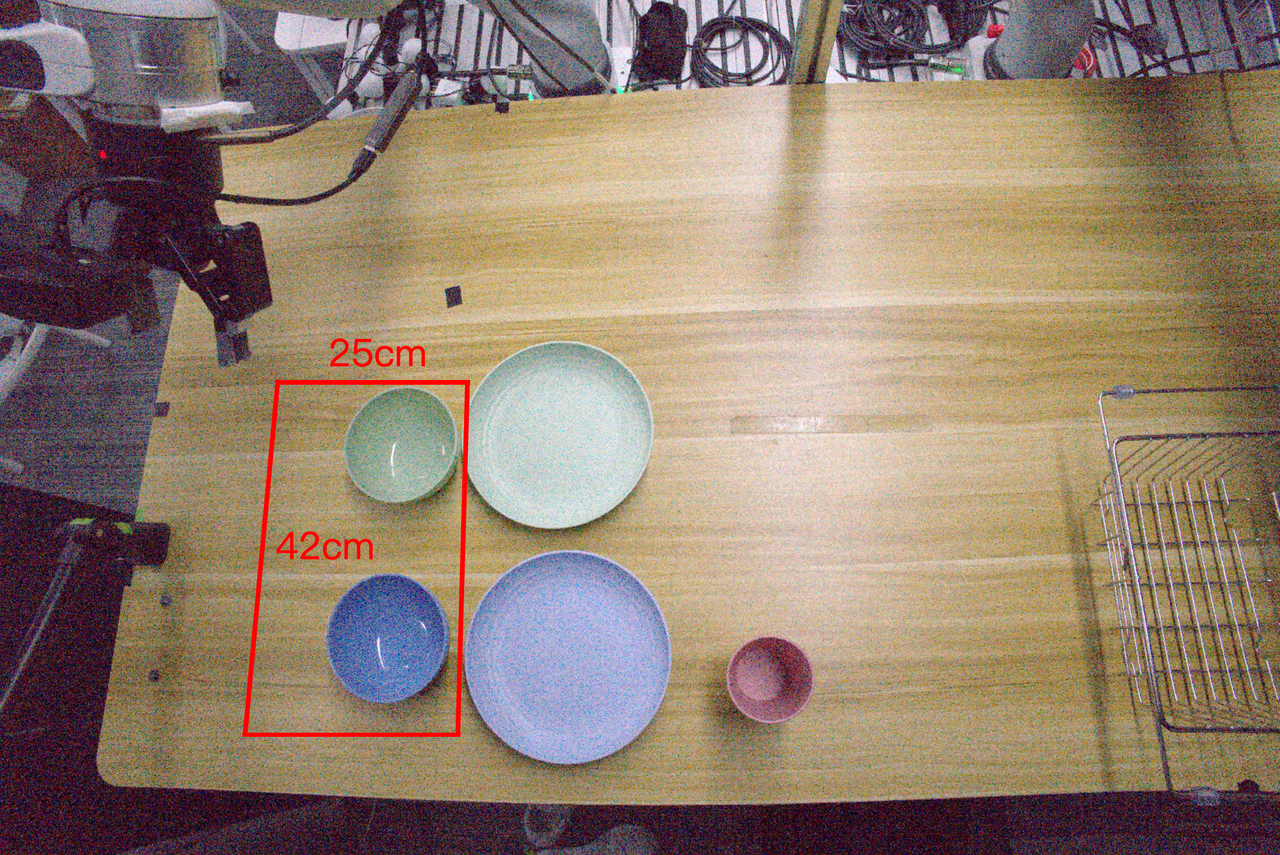}\par\smallskip
  \includegraphics[width=0.49\linewidth]{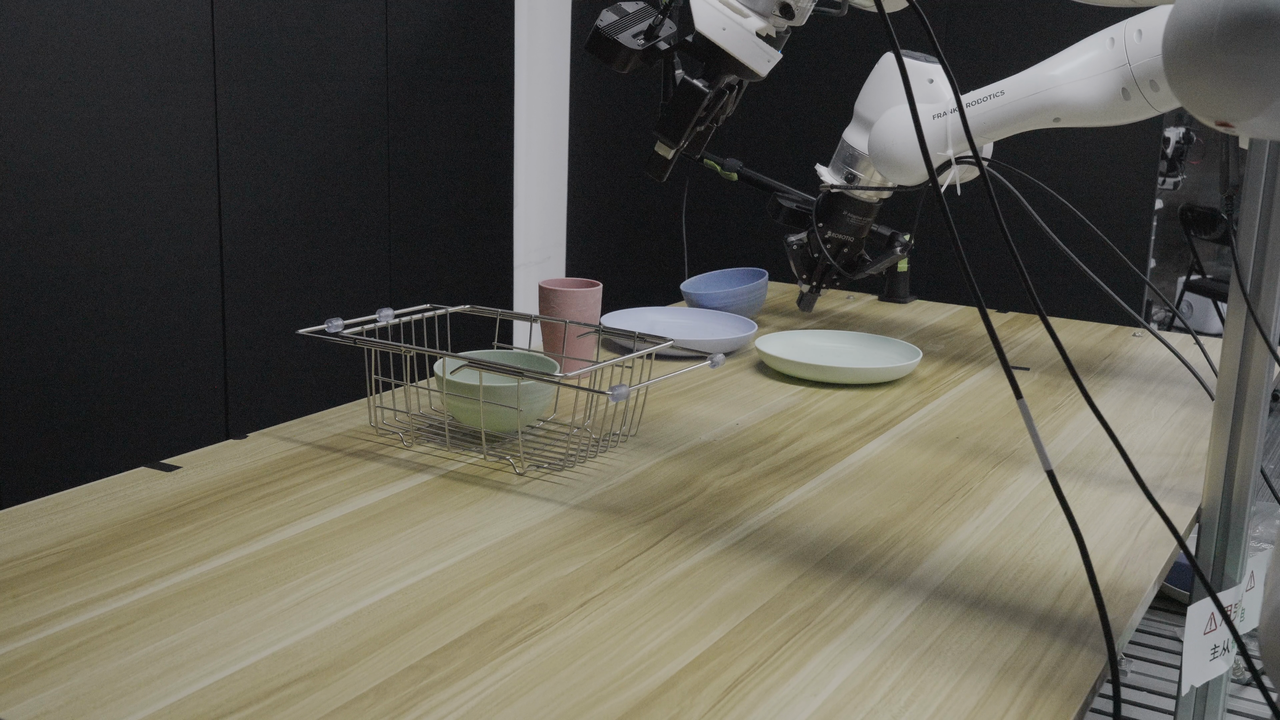}\hfill
  \includegraphics[width=0.49\linewidth]{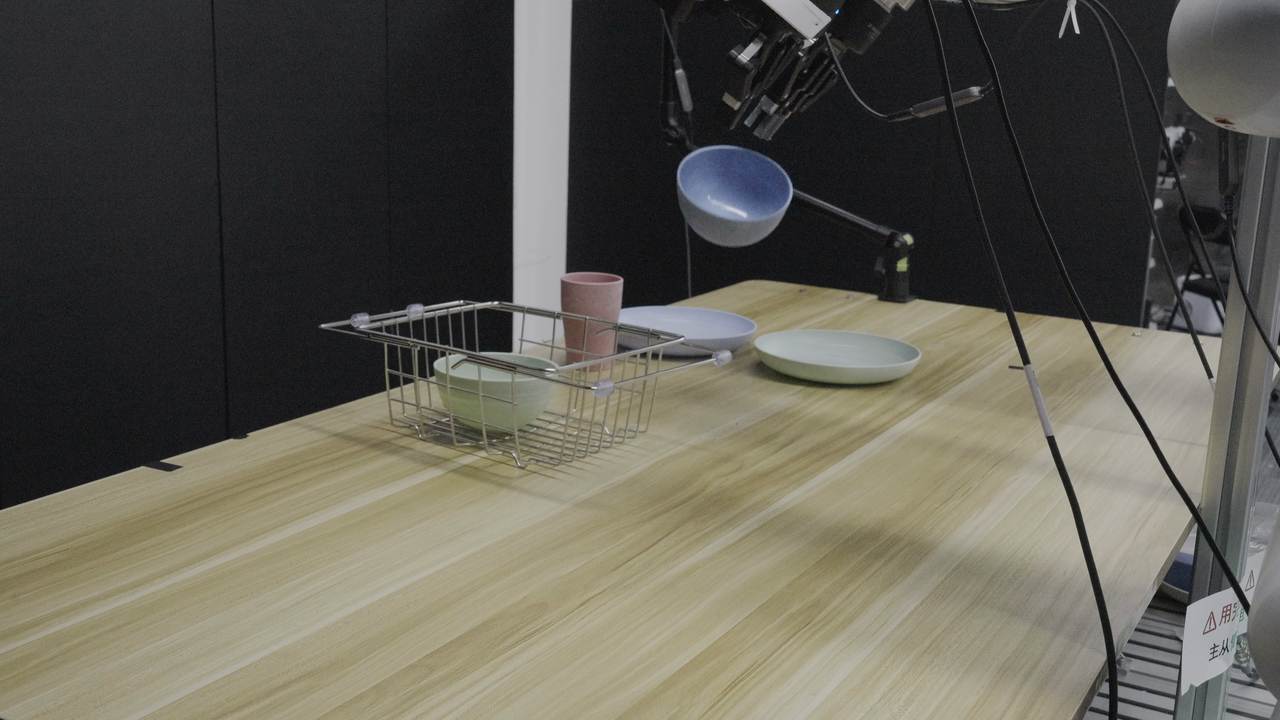}
  \captionof{figure}{Bimanual Table Clearing setup and failures. Top: the red region defines bowl initialization. Bottom left: the right gripper misses the blue bowl's rim. Bottom right: the left gripper fails to secure the bowl during handover, causing it to fall when released.}
  \label{fig:app_table_setup_failures}
\end{center}
\end{appendixtask}

\clearpage
\endgroup
\subsection{Demonstration Scaling and SFT Budgets}
\label{app:human_data_scaling}

Increasing the number of human demonstrations does not necessarily resolve the difficult behaviors in these tasks.
Table~\ref{tab:human_data_scaling} examines whether increasing teleoperation data resolves failures on two difficult long-horizon tasks using human-only SFT, without specialist distillation.
Dual-Arm Peg Insertion remains at 0\% success with 50, 100, 200, and 300 demonstrations: even a sixfold increase yields no successful evaluation episodes under the tested SFT setup.
This task's difficulty therefore persists despite substantial data expansion.
Bimanual Table Clearing benefits from additional data, but with diminishing observed gains: 100, 150, and 300 demonstrations yield 20.0\%, 26.7\%, and 30.0\% success.
The first 50 additional trajectories add 6.7 percentage points, whereas the next 150 add only 3.3 points.

These results motivate keeping demonstration budgets fixed within each task-horizon group instead of continually expanding datasets for difficult tasks.
For SFT initialization in the main experiments, we use 50 human-teleoperated trajectories per short-horizon task and 100 per long-horizon task.
The budget is identical across methods for each task, with subsequent adaptation addressing the remaining precision-critical failures.
Both tasks in Table~\ref{tab:human_data_scaling} are long-horizon tasks; buffer sizes other than 100 trajectories are supplementary scaling controls.

\begin{table}[!ht]
  \centering
  \caption{Human-only SFT scaling on two long-horizon tasks (30 evaluation trials per setting). Trajectory counts refer to human teleoperation demonstrations. The main experiments use 100 demonstrations for each of these tasks; other buffer sizes are supplementary controls. SR is reported in \%; dashes denote untested settings.}
  \label{tab:human_data_scaling}
  \small
  \begin{tabular}{rcc}
    \toprule
    Human trajectories & Dual peg SR $\uparrow$ & Table clearing SR $\uparrow$ \\
    \midrule
    50 & 0.0 & -- \\
    100 & 0.0 & 20.0 \\
    150 & -- & 26.7 \\
    200 & 0.0 & -- \\
    300 & 0.0 & 30.0 \\
    \bottomrule
  \end{tabular}
\end{table}

\end{document}